\documentclass[10pt,twocolumn,letterpaper]{article}
\usepackage[pagenumbers]{cvpr}
\usepackage{amsmath,amssymb,amsthm,booktabs,graphicx,xcolor}
\newcommand{\oursrule}{\specialrule{0.09em}{0.2ex}{0.2ex}}
\usepackage{colortbl}
\usepackage{multirow}
\definecolor{ourspastel}{RGB}{232,245,233}
\definecolor{notepastel}{RGB}{252,248,227}
\definecolor{failpastel}{RGB}{252,233,229}
\usepackage{tikz}
\usetikzlibrary{arrows.meta,calc}
\definecolor{sraA}{RGB}{34,98,175}
\definecolor{sraB}{RGB}{169,74,34}
\definecolor{sraink}{RGB}{58,58,56}
\definecolor{srasoft}{RGB}{185,184,174}
\usepackage{algorithm}
\usepackage{algpseudocode}
\usepackage[pagebackref,breaklinks,colorlinks]{hyperref}
\definecolor{cvprblue}{rgb}{0.21,0.49,0.74}\hypersetup{allcolors=cvprblue}
\usepackage{enumitem}
\usepackage[capitalize]{cleveref}

\usepackage[numbers,sort&compress]{natbib}

\theoremstyle{remark}
\newtheorem{remark}{Remark}

\newcommand{\Prx}{\mathrm{Pr}}
\newcommand{\rr}{\mathbf{r}}
\newcommand{\dd}{\mathbf{d}}
\newcommand{\err}{\mathcal{E}}
\newcommand{\todo}[1]{\textcolor{red}{[#1]}}

\def\paperID{*****}
\def\confName{CVPR}
\def\confYear{2027}

\begin{document}
\raggedbottom
\makeatletter
\renewcommand\paragraph{\@startsection{paragraph}{4}{\z@}{1.5ex \@plus .3ex \@minus .2ex}{-1em}{\normalfont\normalsize\bfseries}}
\makeatother
\makeatletter
\def\endabstract{\par\vspace{12pt}}
\makeatother

\title{Shift--Reliance Alignment: Ranking Vision--Language Models\\
by the Cues They Trust}

\author{Faisal Mohammad$^{1,2}$\qquad Mohamed Younis$^{1}$\\
$^{1}$University of Maryland, Baltimore County\\
$^{2}$Johns Hopkins University Applied Physics Laboratory\\
{\tt\small \{faisalm2,\,younis\}@umbc.edu}}
\maketitle

\begin{abstract}
Vision--language models (VLMs) are usually chosen from leaderboards, which give one ranking for every application. On questions about objects that a scene suggests but does not contain, however, the model with the lowest error on the POPE hallucination benchmark is not the one with the lowest error on such questions. For yes/no questions about object presence, we show that the best model for a target domain can be predicted without running any candidate on that domain, given $650$ to $1{,}200$ labeled target items that calibrate a detector. We measure each model's reliance on a cue, a feature such as scene context or text in the image, by editing $100$ labeled source images so that the cue suggests the wrong answer, and comparing with a matched control edit. Reliance predicts which VLMs are misled by physical sticker attacks (Spearman $.93$ on nine models not used to design the score) and which hallucinate on unedited COCO photographs, beyond what their tendency to answer yes explains. Shift--Reliance Alignment (SRA) scores a model by its source error on the labeled source questions plus its reliance multiplied by the shift. The shift is the proportion of target questions on which the cue misleads, and a detector estimates it on the unlabeled target images. SRA selects the model with the lowest score. In pre-registered tests on new images where most questions mislead, it lowered the error of the selected model from $30.4\%$ under selection by source error to $18.8\%$ on Objects365 and OpenImages but, against its prediction, not on COCO. On CueDial, a benchmark we release that varies how often a cue misleads, selection by model confidence falls more than $40$ percentage points behind the best model when overlaid text denies a visible object, and SRA stays within $3$ points.
\end{abstract}
\begin{figure}[t!]
\centering
\includegraphics[width=0.8\linewidth]{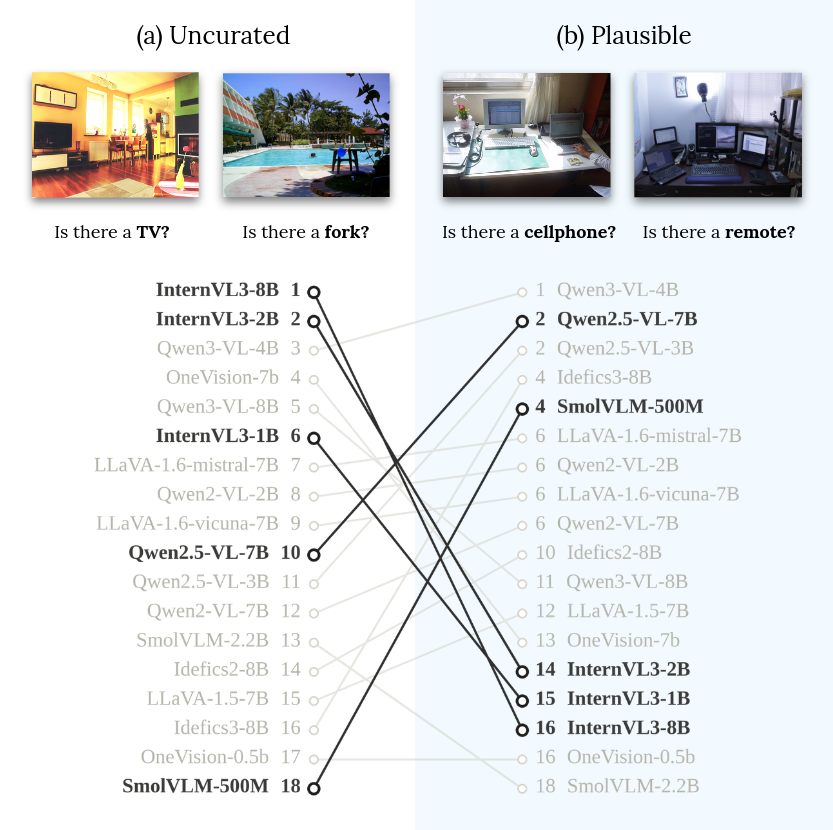}
\caption{\textbf{The same models rank differently on two kinds of test set.} (a) An uncurated test set asks about random objects, here a present TV and an absent fork, and the $18$ models used to develop the score (\cref{sec:setup}) ranked by error on $32$ uncurated test sets. (b) A plausible test set asks only about objects the scene suggests, half of them absent, here a cellphone and a remote that are not there, and the same models ranked by hallucination on one plausible test set, Spearman $-.15$ between the two rankings.}
\label{fig:teaser}
\end{figure}
\section{Introduction}
\begin{figure*}[t]
\centering
\includegraphics[width=0.8\textwidth]{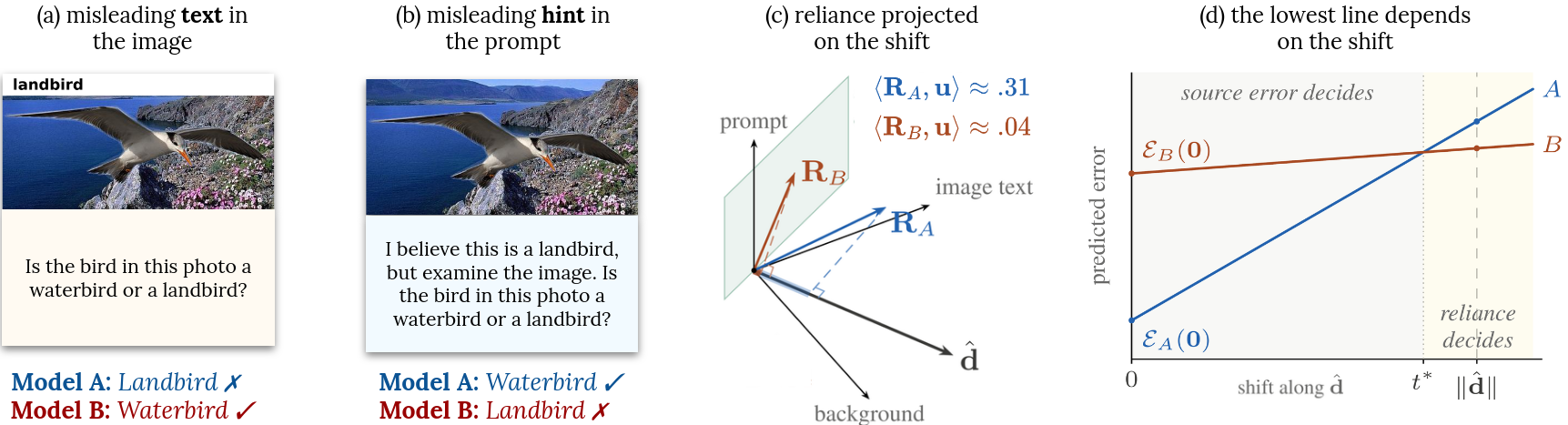}
\caption{\textbf{How SRA chooses between two models} on Waterbirds~\cite{sagawa2020groupdro}, Qwen2.5-VL-3B ($A$) and LLaVA-1.5-7B ($B$). (a,\,b) Misleading text changes $A$'s answer but not $B$'s, and a misleading hint in the prompt changes $B$'s answer but not $A$'s. (c) The reliance of each model and the shift $\hat{\mathbf d}=(1,.5,0)$ of a target domain over background, image text, and prompt. Each unit of shift along $\hat{\mathbf d}$, with unit direction $\mathbf u$, adds $.31$ error to $A$ and $.04$ to $B$. (d) Predicted error grows from each source error along $\hat{\mathbf d}$, the two lines cross at $t^{*}$, and no single ranking is right on both sides of it. SRA selects $B$, whose realized error is lower ($.488$ against $.538$, \cref{app:wbdetail}).}
\label{fig:formulation}
\end{figure*}

Leaderboards~\cite{duan2024vlmevalkit} are the usual way to choose a vision--language model (VLM), but a leaderboard gives one ranking for every application, and which models fail differs from one application to the next. Consider a model that checks whether expected objects are present, such as safety equipment in a building inspection. Many of its questions ask about objects the scene suggests, and when one is missing, the scene itself suggests the wrong answer. Across $18$ models, the ranking by error on questions about random objects is nearly unrelated to the ranking by hallucination on questions about objects the scene suggests (\cref{fig:teaser}), and on test sets where most questions ask about such objects, the model with the lowest error on the POPE hallucination benchmark~\cite{li2023pope} trails the best model by $3$ to $21$ percentage points on average, depending on the corpus (\cref{sec:selection}). Labels for the target images are scarce, and running every candidate on the target is costly when candidates and targets are many. Unsupervised accuracy estimators~\cite{garg2022atc,deng2023dispersity,xie2024mano} need no labels but run every candidate on the unlabeled target images and rely on each model's confidence, and a misled model often remains confident (\cref{sec:dialresults}). This raises two questions. Can a model's failures on a target domain be predicted before the model is run there? And can such predictions select the model with the lowest error on that domain, the problem of model selection under distribution shift~\cite{sugiyama2007iwcv,you2019dev,saito2021snd}? We call any rule that picks one model per test set a selector.

\begin{figure*}[t]
\centering
\includegraphics[width=\textwidth]{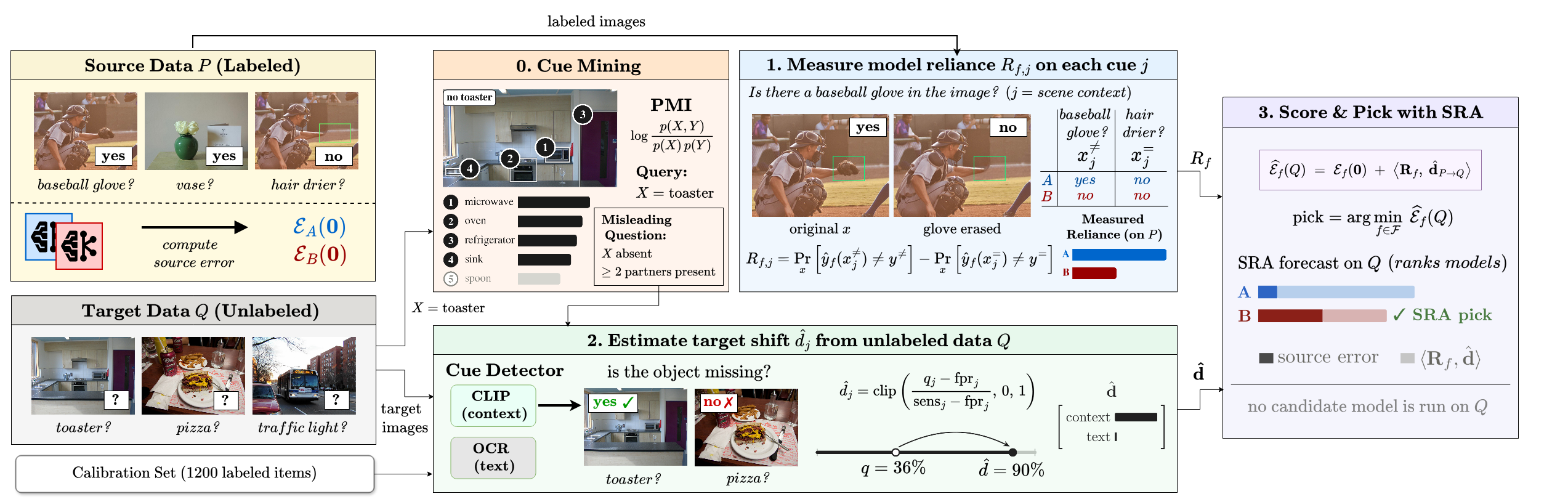}
\caption{\textbf{The SRA pipeline on COCO test sets of mostly misleading questions.} (0) Partners, the categories most associated with the queried object, are mined from COCO and define the misleading questions (\cref{sec:discovered}). (1) Reliance is a candidate's error on labeled source images after the object is erased, minus its error on an unrelated absent object in the same image (\cref{sec:relmeas}). (2) A detector flags target questions whose object appears missing, and its error rates, measured on labeled items disjoint from the test sets, correct the flagged proportion (\cref{sec:dhat}). The marks \checkmark{} and $\times$ show whether the detector is right, which the method never sees. (3) SRA selects $B$ although $A$ has the lower source error (\cref{eq:sra}).}
\label{fig:hero}
\end{figure*}

We answer both with one idea. A cue is an input feature, such as scene context or text in the image, that the robustness literature calls a spurious feature~\cite{geirhos2020shortcut}. We score a model for a target domain by its source error, its error on labeled source questions, plus, for each cue, the extra error that cue causes, multiplied by how often the cue misleads on the target. Source error and the extra error, the model's reliance on the cue, are measured once on source images, reliance by editing that cue and subtracting the error under a matched control edit. A detector estimates the proportion of misleading questions for each cue on the unlabeled target images, forming the shift vector $\hat\dd$, and its error rates are measured on a labeled sample of unedited photographs. This sum, source error plus the inner product of reliance and shift, is the Shift--Reliance Alignment (SRA) score, and we select the model with the lowest one (\cref{fig:formulation,fig:hero}). Each candidate runs only on source images while its rank still changes with the target. Evaluating such a selector requires test sets whose proportion of misleading questions is known and varied, so we build CueDial.

\paragraph{Contributions.}
\begin{itemize}[leftmargin=1.2em,itemsep=1pt,topsep=2pt]
\item \textbf{SRA.} We propose a selection score that ranks VLMs for a target domain by their source error, their reliance on each cue, measured once on $100$ edited source images, and the estimated shift. It runs no candidate on the target, and its only target labels are the $650$ to $1{,}200$ items that calibrate a cue detector once. Reliance alone predicts which VLMs fail on existing benchmarks, from physical stickers to hallucination on unedited COCO photographs (\cref{sec:transfer}), and in a pre-registered test on new Objects365 and OpenImages images where $80\%$ of questions mislead, SRA lowered the error of the selected model by more than a third (\cref{sec:selection}).
\item \textbf{CueDial.} We release a benchmark that varies the proportion of misleading questions from $0$ to $.5$ for two cue types over six labeled datasets, with the cached answers of all $40$ models, audited labels, and a hidden split (\cref{sec:cuedial,sec:dialresults}).
\end{itemize}

\section{Related Work}\label{sec:related}

\paragraph{Unsupervised accuracy estimation and model selection.}
Several methods estimate a model's accuracy on an unlabeled target set under distribution shift. Confidence-based methods include Average Thresholded Confidence (ATC)~\cite{garg2022atc} and the difference of confidences (DoC)~\cite{guillory2021predicting}. Other methods use the agreement between models~\cite{baek2022agreement,jiang2022disagreement,saxena2024foundation,miller2021accuracyontheline}, distribution distances~\cite{deng2021autoeval}, the nuclear norm of the prediction matrix~\cite{deng2023dispersity}, or the matrix norm of normalized logits (MaNo)~\cite{xie2024mano}, and Ranked from Within~\cite{tu2025ranked} ranks large multimodal models by the probabilities of their answers. All need every candidate's outputs on the target, while SRA never runs a candidate there, and ATC and confidence optimal transport (COT)~\cite{lu2023cot} also use labeled source data. The closest method is Mandoline~\cite{chen2021mandoline}, which reweights labeled source data along slices that the practitioner specifies. LOVM~\cite{zohar2023lovm}, VEGA~\cite{ding2024vega}, and OSTB~\cite{xu2026ostb} rank CLIP-style models by a task description or a shared embedding, which generative VLMs lack.

\paragraph{Measuring VLM cue failures.}
Hallucination benchmarks each fix the proportion of misleading questions by construction~\cite{rohrbach2018chair,li2023pope,wang2023amber,guan2024hallusionbench,augustin2025dash}, while MetaShift~\cite{liang2022metashift} controls context shift for image classifiers. Hallucination has been traced to language priors~\cite{leng2024vcd,liu2024hallusurvey} and object co-occurrence~\cite{zhou2024lure}, and context shapes object recognition more broadly~\cite{zhang2020context}. Typographic attacks exploit sensitivity to text in the image~\cite{goh2021multimodal,pixels2026typographic}. AutoHallusion~\cite{wu2024autohallusion} generates images whose context invites hallucination, Causal-HalBench~\cite{xu2026causalhalbench} swaps in objects that rarely co-occur, and He et al.~\cite{he2025removed} erase objects by inpainting, each reporting a per-model failure rate close to our context reliance, whereas we subtract the error under a matched control and use the result to select models. Counterfactual edits have probed one model at a time~\cite{veitch2021counterfactual,lang2021stylex,prabhu2023lance,vendrow2023dataset}, and methods that discover error slices or biases~\cite{eyuboglu2022domino,kim2024b2t} can supply candidate cues. Imperfect edits can bias such estimates~\cite{reber2024rate,taufiq2023manifold}, which the matched control and placebo edits check for our inpainting edits~\cite{suvorov2022lama,telea2004inpainting}.

\section{Method}\label{sec:law}
\begin{table*}[t]
\centering\footnotesize
\caption{\textbf{CueDial against related benchmarks.} Only CueDial sets the proportion of misleading questions by design, releases every model's answer on every item, and scores selectors across that proportion. $\circ$ marks partial support, since SCAM pairs each sticker photo with a clean copy, POPE's adversarial split raises the proportion without setting it, and LOVM releases accuracies, not answers.}
\label{tab:cuedialcompare}
\setlength{\tabcolsep}{2.2pt}
\newcommand{\y}{\checkmark}\newcommand{\n}{--}\newcommand{\p}{$\circ$}
\begin{tabular}{l>{\columncolor{ourspastel}}ccccccc}
\toprule
& \textbf{CueDial} & POPE~\cite{li2023pope} & AMBER~\cite{wang2023amber} & DASH-B~\cite{augustin2025dash} & SCAM~\cite{westerhoff2025scam} & \shortstack{Ranked from\\Within~\cite{tu2025ranked}} & LOVM~\cite{zohar2023lovm}\\
\midrule
\shortstack[l]{Controlled proportion of\\misleading questions} & \y & \p & \n & \n & \p & \n & \n\\
Cue types & text, context & context & context & context & text & -- & --\\
Source datasets & 6 & 3 & 1 & web images & 1 & 9 (visual question answering) & 23\\
Answer formats & yes/no & yes/no & yes/no, free text & yes/no & two-way choice & choice, open & classification\\
Model answers per item & \y{} ($40$ models) & \n & \n & \n & \n & \n & \p\\
Scores & selectors & models & models & models & models & rankers & rankers, estimators\\
Hidden test split & \y & \n & \n & \n & \n & \n & \n\\
\bottomrule
\end{tabular}
\end{table*}

\subsection{Shift--Reliance Alignment}\label{sec:sra}
Consider candidate models $f$ that answer yes/no questions about images, a labeled source set from a distribution $P$, and unlabeled images from a target domain $Q$, with no target labels beyond a sample used only to measure a detector's error rates (\cref{sec:dhat}). Cues $c_1,\dots,c_k$ can suggest the wrong answer to a question, and we say that a cue misleads on such a question. SRA scores each model on $Q$ by
\begin{equation}
\widehat{\err}_f(Q)=\err_f(\mathbf 0)+
\big\langle \mathbf{R}_f,\ \hat\dd_{P\to Q}\big\rangle
\label{eq:sra}
\end{equation}
and selects the model with the lowest score. The source error $\err_f(\mathbf 0)$, the score at zero shift, is the model's mean error on $200$ labeled source questions. Half ask about objects present in unedited COCO images, and half ask about unrelated absent objects on the edited images of \cref{sec:relmeas}, where the same questions serve as the control for scene context. A model's mean probability of answering yes on these absent questions is its answer bias, akin to the yes ratio of POPE~\cite{li2023pope}, and counting these questions keeps source error from favoring a model that answers yes to everything (\cref{sec:transfer}). The reliance $\mathbf R_f\in\mathbb R^k$ is the extra error the model makes when each cue misleads (\cref{sec:relmeas}), and the shift $\hat\dd_{P\to Q}\in[0,1]^k$ holds the estimated proportion of misleading questions for each cue in $Q$ (\cref{sec:dhat}). Only the inner product in \cref{eq:sra}, the reliance term, depends on $Q$, so each model is measured once, on the source, while its rank can change with the target (\cref{app:scalar}). With a single cue, the target error is linear in the shift if each model's error is constant over the misleading questions and over the rest (\cref{app:theory}).

\subsection{Cues}\label{sec:discovered}
Cues can be named by hand, as in \cref{fig:formulation}, or mined from annotations. For each object category $X$ in a corpus, we rank the other categories by pointwise mutual information (PMI) with $X$~\cite{church1990pmi} and call the top five the partners of $X$, akin to the co-occurring objects that POPE's adversarial split asks about~\cite{li2023pope}. The scene context cue is the presence of these partners, and a question about an absent $X$ is a misleading question when at least two of its partners are present (\cref{fig:hero}, step 0).

\subsection{Measuring reliance \texorpdfstring{$\mathbf R_f$}{R}}\label{sec:relmeas}
We measure reliance on cue $j$ by editing the labeled source images so that the cue suggests the wrong answer and subtracting the error under a matched control edit that does not mislead (\cref{fig:hero}, step 1),
\begin{equation}
R_{fj}=\Prx_x\big[\hat y_f(x^{\neq}_j)\neq y^{\neq}\big]-\Prx_x\big[\hat y_f(x^{=}_j)\neq y^{=}\big],
\label{eq:flip}
\end{equation}
where $x^{\neq}_j=\Phi^{\neq}_j(x)$ and $x^{=}_j=\Phi^{=}_j(x)$ are the images after the misleading and the control edits, $y^{\neq}$ and $y^{=}$ their correct answers, and in practice each indicator is the model's probability of the wrong answer. For scene context, the edit erases every instance of $X$ by inpainting~\cite{suvorov2022lama} while its partners stay, and the control keeps the same edited image but asks about an unrelated absent object. For text we use two edits. The overlay edit pastes a sentence that denies the object, with the error averaged over four templates (\cref{app:reversaldetail}), and its control pastes a sentence that agrees. The sticker edit pastes a word naming an absent object, as sticker attacks do, and its control pastes a word naming the present object (\cref{sec:transfer}). On two models, placebo edits move the answers by about a tenth of what erasing the object does (\cref{app:artifacts}).

\begin{figure}[t]
\centering
\includegraphics[width=\linewidth]{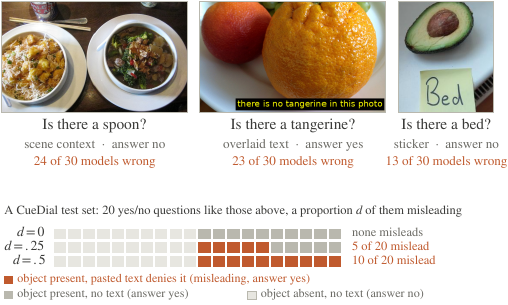}
\caption{\textbf{Three CueDial questions from three families}, with how many of the $30$ candidates answer each wrongly. Below, a test set for the text cue at $d=0$, $.25$, and $.5$, shown with $20$ questions. Half the answers are yes.}
\label{fig:cuedialsamples}
\end{figure}
\subsection{Estimating the shift \texorpdfstring{$\hat\dd_{P\to Q}$}{d}}\label{sec:dhat}
Each coordinate of $\hat\dd$ comes from a cue detector $g_j$ outside the candidate set. Optical character recognition (OCR) with EasyOCR~\cite{easyocr} detects misleading text. For scene context, on questions whose partners are present, CLIP~\cite{radford2021clip} scores the queried object's absence on COCO, and on OpenImages and Objects365 an object detector scores absence and CLIP the partners (\cref{app:procedure}). A labeled sample from the same corpus, the calibration set, serves only to measure the detector's error rates, its sensitivity $\mathrm{sens}_j$ and false-positive rate $\mathrm{fpr}_j$. We then correct the flagged proportion $q_j$ with the standard correction for an imperfect test~\cite{rogan1978prevalence}, $\hat d_j=\mathrm{clip}\big((q_j-\mathrm{fpr}_j)/(\mathrm{sens}_j-\mathrm{fpr}_j),0,1\big)$. On OpenImages and Objects365 the detector gives two signals per question. We model their counts on the target as a mixture of three question types (misleading, present, and absent objects unrelated to the scene), measure each type's signal rates on the calibration set, and estimate the three proportions by nonnegative least squares (\cref{app:newcorpora}).

\subsection{CueDial}\label{sec:cuedial}
CueDial varies the proportion of misleading questions by design (\cref{fig:cuedialsamples,tab:cuedialcompare}). Each family pairs one cue $j$ with one image dataset and draws from three pools. In the misleading pool $\mathcal C$ the cue contradicts the correct answer, in the neutral pool $\mathcal N$ the answer is the same as in $\mathcal C$ but no cue misleads, and in the third pool $\mathcal O$ the answer is the opposite one and no cue misleads. A test set of $n$ questions in which a proportion $d\in[0,\frac12]$ mislead draws $T_d=\mathcal O_{n/2}\cup\mathcal C_{dn}\cup\mathcal N_{(1/2-d)n}$, where a subscript gives the number of items drawn without replacement, so the yes/no balance stays at one half and only the proportion of misleading questions changes. By construction, each model's expected error is linear in $d$,
\begin{equation}
\mathbb E\big[\err_f(T_d)\big]=\tfrac12\,e_f^{\mathcal O}+\tfrac12\,e_f^{\mathcal N}+d\,\big(e_f^{\mathcal C}-e_f^{\mathcal N}\big),
\label{eq:dialerr}
\end{equation}
where $e_f^{\mathcal X}$ is the model's error on pool $\mathcal X$, so two models' lines cross at most once, the best model changes only at such crossings, and the SRA score has the same linear form, with source error as the intercept. A selector's regret on a test set is the error of the model it selects minus the error of the best candidate, so zero regret means the best model was chosen. We summarize a selector $s$ on a family by its worst-case mean regret $W_s$, its mean regret over the test sets drawn at a proportion $d$, maximized over $d$,
\begin{equation}
W_s=\max_d \mathbb E\big[\err_{s(T_d)}(T_d)-\min_f\err_f(T_d)\big].
\label{eq:ws}
\end{equation}
We draw $30$ test sets of $n=200$ questions at each of $11$ values of $d$ between $0$ and $.5$. The eight families, four per cue, pair scene context with COCO, LVIS, OpenImages, and Objects365, and text with sentences pasted on the AMBER and COCO images of the constructed test sets, SCAM's physical stickers, and SynthSCAM, in which the sticker word is pasted digitally on SCAM's photographs, all drawn from the images of the test sets of \cref{sec:setup}. Claude Opus 5.5, a multimodal model outside the candidate set, checked the items of the test-set types of \cref{tab:cuedialaudit} on which at least $27$ of the $30$ candidates disagree with the label, and we dropped the $835$ items it flagged (\cref{app:cuedial}).

\section{Experiments}\label{sec:exp}
\begin{figure*}[t]
\centering
\begin{minipage}[c]{0.67\textwidth}\centering
\includegraphics[width=\linewidth]{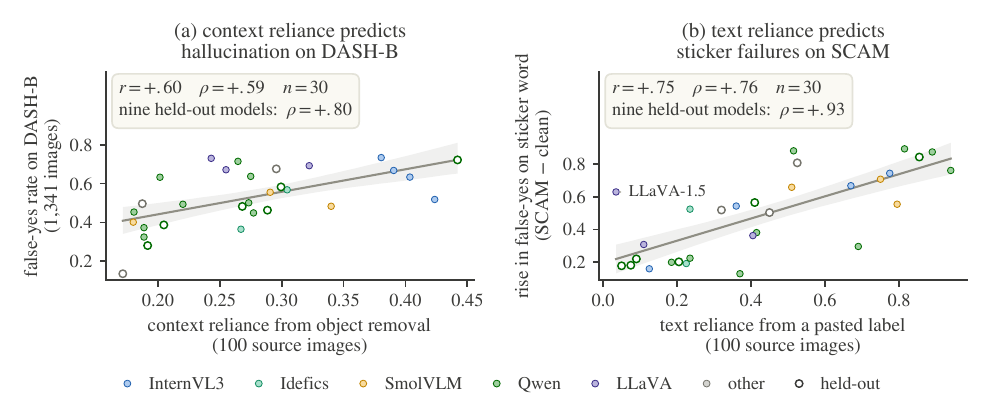}
\end{minipage}\hfill
\begin{minipage}[c]{0.31\textwidth}
\caption{\textbf{Reliance from source edits predicts failure on existing benchmarks.} (a) DASH-B's $1{,}341$ images without the queried object. (b) SCAM's physical stickers. Hollow points are the nine held-out models (\cref{sec:setup}), and LLaVA-1.5 cannot read the pasted label.}
\label{fig:dayzero}
\end{minipage}
\end{figure*}
\subsection{Setup}\label{sec:setup}
\begin{table}[t]
\centering\scriptsize
\caption{\textbf{The $87$ test sets} and the proportion of their questions that mislead. $2{+}3$ means misleading plus ordinary test sets, the latter about random objects of the same corpus, $4{+}2$ plausible plus ordinary ones, and -- a proportion not measured. CueDial's test sets, the $20$ pre-registered test sets, and the six calibrated DASH-B test sets of \cref{app:organic} are not among the $87$ (\cref{app:cuedial,app:confirm}).}
\label{tab:testsets}
\setlength{\tabcolsep}{1.5pt}
\begin{tabular}{@{}lc p{4.3cm} c@{}}
\toprule
type & sets & images and questions & misleading\\
\midrule
uncurated & $32$ & random objects, four corpora & ${\approx}.02$\\
constructed & $10$ & edited AMBER and COCO images & $.50$\\
COCO misleading & $4$ & train2017, absent, partners present & $.80$\\
OpenImages & $2{+}3$ & validation split, misleading and ordinary & $.80$, ${<}.05$\\
Objects365 & $4{+}3$ & one train shard, misleading and ordinary & $.80$, ${\le}.01$\\
SCAM & $6$ & sticker photos, clean and pasted copies & varies\\
plausible & $6$ & COCO, LVIS, scene suggests, \mbox{half absent} & $.50$\\
mixed & $4$ & plausible, half random absent objects & $.25$\\
POPE & $3$ & its three splits & $.01$--$.09$\\
VQAv2 & $2$ & human questions about COCO objects & $.07$\\
DASH-B & $2$ & mined web images, \mbox{half absent} & ${\approx}.5$\\
Objects365 plausible & $4{+}2$ & plausible and random objects, other images & $.51$--$.54$, --\\
\bottomrule
\end{tabular}
\end{table}
\paragraph{Models.} We evaluate $40$ open VLMs~\cite{zhu2025internvl3,wang2024qwen2vl,bai2025qwen25vl,bai2025qwen3vl,liu2023llava,liu2024llava15,liu2024llavanext,li2024llavaonevision,laurencon2024idefics2,laurencon2024idefics3,marafioti2025smolvlm}, $30$ of which form the candidate pool, on questions of the form ``is there a [object]?'' (\cref{app:impl}). We designed the score on $18$ of the $30$, the development models, and held out the remaining $12$. Nine held-out models are reported separately in the correlations of \cref{sec:transfer}, and the other three join the development models as the $21$ selection candidates. The DASH-B and SCAM correlations and CueDial use all $30$, the COCO hallucination correlation the $18$ development models, and the free-text naming test, in which models name the object, all $40$. Half of the $30$ candidates run with 4-bit quantization.

\paragraph{Test sets.} \Cref{tab:testsets} lists the $87$ test sets used for selection. Uncurated test sets ask about random objects, constructed ones use images we edited, plausible ones ask only about objects the scene suggests, half of them absent, and misleading ones ask mostly about absent objects whose partners are present. They are drawn from COCO~\cite{lin2014coco}, GQA~\cite{hudson2019gqa}, LVIS~\cite{gupta2019lvis}, COCO-Text~\cite{veit2016cocotext}, VQAv2~\cite{goyal2017vqav2}, OpenImages~\cite{kuznetsova2020openimages}, Objects365~\cite{shao2019objects365}, AMBER~\cite{wang2023amber} (a hallucination benchmark whose images we edit), SCAM~\cite{westerhoff2025scam} (photographs of objects with a handwritten sticker naming another object), and DASH-B~\cite{augustin2025dash} (web images mined to make VLMs hallucinate). \Cref{sec:transfer} also uses the $1{,}341$ DASH-B images without the queried object and the COCO hallucination set, $558$ questions about absent objects the scene suggests on unedited COCO images.

\begin{figure*}[t]
\centering
\begin{minipage}[c]{0.60\textwidth}\centering
\includegraphics[width=\linewidth]{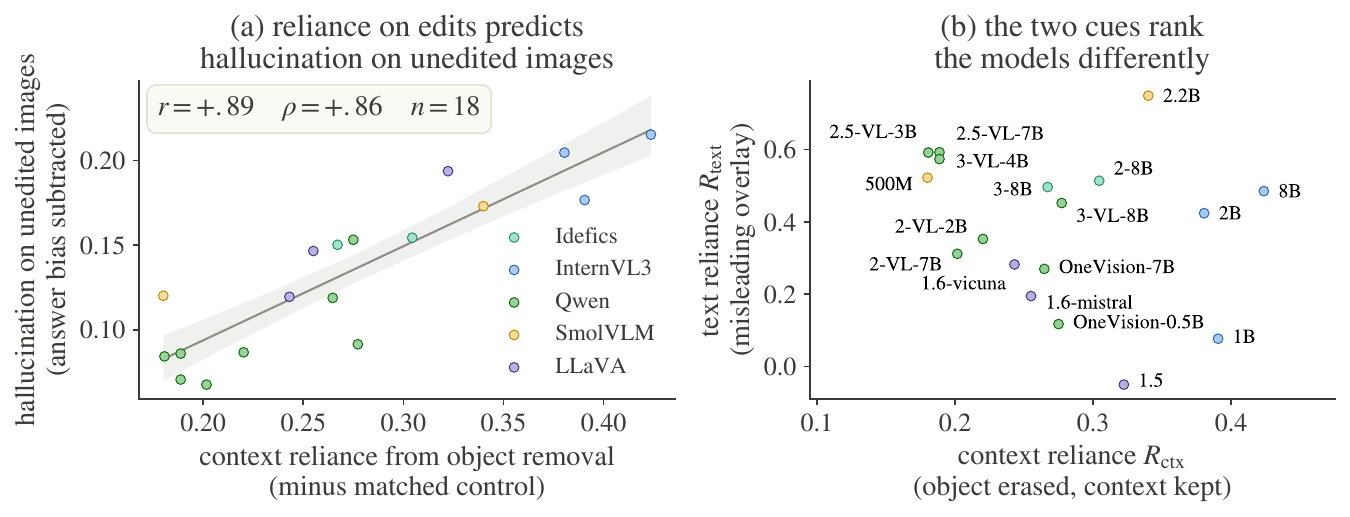}
\end{minipage}\hfill
\begin{minipage}[c]{0.38\textwidth}
\caption{\textbf{Reliance measured on edits predicts hallucination on unedited images, and reliance on one cue says little about reliance on the other.} (a) Hallucination on the COCO hallucination set with each model's answer bias subtracted, $18$ VLMs, with a least-squares fit and bootstrap band. (b) Text reliance $R_{\mathrm{text}}$ on overlays that deny the object against context reliance $R_{\mathrm{ctx}}$ for the same models, colored as in (a) with OneVision under Qwen, its backbone (Spearman $-.34$, $95\%$ confidence interval (CI) $[-.76,.20]$).}
\label{fig:transferplane}
\end{minipage}
\end{figure*}
\paragraph{Protocol.} The reliance predictions for the held-out models, the sticker correlations, the DASH-B Pearson correlation over all $30$ models, and the tests on new images (\cref{sec:hc,app:confirm}) were fixed as SHA-256 hashes before the models concerned answered any test item.

\paragraph{Baselines.} Selection by source error picks the model with the lowest source error. Label-shift reweighting weights each model's source error on yes and no questions by the test set's true yes/no balance, an oracle version of black-box label-shift estimation~\cite{lipton2018bbse}. Selection by POPE error picks the model with the lowest error on POPE, and selection by least answer bias the model with the lowest answer bias (\cref{sec:sra}). The fixed POPE choice picks for every test set the model with the lowest false ``yes'' rate on POPE's adversarial split. The best single model, Qwen3-VL-4B, is the one with the lowest regret in hindsight, a reference rather than a selector. We also compare against nine unsupervised accuracy estimators, which require each candidate's outputs on the test set. They are ATC~\cite{garg2022atc}, DoC~\cite{guillory2021predicting}, average confidence, entropy, the nuclear norm~\cite{deng2023dispersity}, two variants of MaNo~\cite{xie2024mano}, model agreement~\cite{baek2022agreement}, and majority-vote pseudo-labels (\cref{sec:related,app:lfall}). \Cref{tab:calconf,tab:hc} report them raw, and \cref{tab:calconffull} also after Platt scaling~\cite{platt1999probabilistic}.

\paragraph{Metrics.} Regret (\cref{sec:cuedial}) is in percentage points here and in fractions in the appendix unless marked as points. Correlations are Spearman unless marked $r$ (Pearson). In tables, our rows are bold, green marks the best value in a column within a block, yellow a value in our rows that is neither best nor poor, and red the failure named in each caption, which for regret is a value above a random choice's and takes precedence over green.

\subsection{Reliance predicts failures on existing benchmarks}\label{sec:transfer}

\paragraph{Physical stickers.} Reliance on a pasted word predicts which models SCAM's physical stickers mislead. The stickers carry a handwritten word naming an absent object, and we measure reliance on such a word by pasting it onto the source images. This reliance correlates with the rise in a model's false ``yes'' rate under the real stickers, both over the $30$ models and over the nine held-out ones (\cref{fig:dayzero}b, \cref{tab:pred}). In free text it ranks the ten models outside the $30$ candidates by the stickers' effect on naming (\cref{app:scamopen}).

\begin{table}[t]
\centering\footnotesize
\caption{\textbf{Reliance predicts hallucination without using the test images.} Spearman correlation between a per-model score and the hallucination rate on the COCO hallucination set ($18$ development models), the false ``yes'' rate on DASH-B, or the rise in false ``yes'' answers under SCAM's stickers. Reliance is context reliance, and on SCAM reliance on a pasted word naming another object. The partial row holds answer bias fixed. Error on present objects correlates negatively because it rewards models that answer yes readily. Partials over nine models are fragile. Red marks a correlation at or below zero.}
\label{tab:pred}
\setlength{\tabcolsep}{2.2pt}\scriptsize
\begin{tabular}{lccccc}
\toprule
 & COCO\,$\uparrow$ & \multicolumn{2}{c}{DASH-B\,$\uparrow$} & \multicolumn{2}{c}{SCAM stickers\,$\uparrow$}\\
score & halluc.\ ($18$) & all $30$ & $9$ held-out & all $30$ & $9$ held-out\\
\midrule
\multicolumn{6}{l}{\emph{never runs the model on the test images}}\\
answer bias & $+.53$ & \cellcolor{ourspastel}$+.70$ & \cellcolor{ourspastel}$+.98$ & $+.36$ & $+.55$\\
error on present objects & \cellcolor{failpastel}$-.55$ & \cellcolor{failpastel}$-.52$ & \cellcolor{failpastel}$-.92$ & \cellcolor{failpastel}$-.08$ & \cellcolor{failpastel}$-.42$\\
\textbf{reliance $R_f$ (ours)} & \cellcolor{ourspastel}$+.78$ & \cellcolor{notepastel}$+.59$ & \cellcolor{notepastel}$+.80$ & \cellcolor{ourspastel}$+.76$ & \cellcolor{ourspastel}$+.93$\\
\textbf{$R_f$, partial (ours)} & \cellcolor{notepastel}$+.77$ & \cellcolor{notepastel}$+.47$ & \cellcolor{failpastel}$-.03$ & \cellcolor{notepastel}$+.73$ & \cellcolor{notepastel}$+.90$\\
\midrule
\multicolumn{6}{l}{\emph{reference: uses each model's answers on the test images}}\\
ATC & \cellcolor{ourspastel}$+.88$ & $+.38$ & \cellcolor{ourspastel}$+.72$ & $+.43$ & $+.43$\\
DoC & $+.81$ & \cellcolor{ourspastel}$+.52$ & $+.65$ & $+.36$ & $+.08$\\
average confidence & $+.85$ & $+.39$ & $+.15$ & $+.24$ & \cellcolor{failpastel}$-.08$\\
entropy & $+.82$ & $+.39$ & $+.15$ & $+.20$ & \cellcolor{failpastel}$-.25$\\
nuclear norm & $+.20$ & $+.38$ & \cellcolor{failpastel}$-.17$ & \cellcolor{ourspastel}$+.70$ & \cellcolor{ourspastel}$+.72$\\
MaNo & $+.85$ & $+.39$ & $+.15$ & $+.23$ & \cellcolor{failpastel}$-.08$\\
\bottomrule
\end{tabular}
\end{table}
\paragraph{Hallucination.} Context reliance predicts hallucination on unedited COCO photographs beyond what answer bias explains. It ranks the $18$ development models by hallucination rate on the COCO hallucination set (\cref{tab:pred}, \cref{app:ablate,app:pred}), more closely once answer bias is subtracted (\cref{fig:transferplane}). With answer bias held fixed, reliance still predicts hallucination and SCAM failures, while on DASH-B (\cref{fig:dayzero}a) answer bias itself predicts better (\cref{tab:pred}). On SCAM and DASH-B reliance correlates with failure more strongly than the six estimators of \cref{tab:pred}, and on the COCO hallucination set five of the six do better (\cref{tab:pred}).

\begin{table}[t]
\centering\scriptsize
\caption{\textbf{SRA has the lowest regret on the constructed test sets.} Mean regret in points, $21$ candidates, over the $32$ uncurated, $10$ constructed, and $6$ plausible test sets of \cref{tab:testsets} and over the misleading test sets of each corpus. The reference block runs every candidate on the test set, and the best single model, Qwen3-VL-4B, is never green. \Cref{tab:calconffull} adds the mixed and POPE test sets.}
\label{tab:calconf}
\setlength{\tabcolsep}{1pt}
\begin{tabular*}{\linewidth}{@{\extracolsep{\fill}}lcccccc}
\toprule
 & \multicolumn{3}{c}{test-set type} & \multicolumn{3}{c}{misleading test sets}\\
\cmidrule(lr){2-4}\cmidrule(lr){5-7}
selector & uncurated & constructed & plausible & COCO & \shortstack{Open-\\Images} & Objects365\\
\midrule
source error & $1.6$ & \cellcolor{failpastel}$10.9$ & \cellcolor{failpastel}$5.7$ & \cellcolor{ourspastel}$2.3$ & \cellcolor{failpastel}$13.6$ & \cellcolor{failpastel}$12.5$\\
label-shift reweighted & $1.6$ & \cellcolor{failpastel}$10.9$ & \cellcolor{failpastel}$5.7$ & \cellcolor{failpastel}$4.9$ & \cellcolor{failpastel}$21.2$ & \cellcolor{failpastel}$13.0$\\
POPE error & \cellcolor{ourspastel}$1.3$ & \cellcolor{failpastel}$14.4$ & \cellcolor{failpastel}$7.2$ & $3.4$ & \cellcolor{failpastel}$21.2$ & \cellcolor{failpastel}$13.4$\\
fixed POPE choice & \cellcolor{failpastel}$4.5$ & \cellcolor{failpastel}$13.2$ & \cellcolor{ourspastel}$2.2$ & $2.5$ & \cellcolor{ourspastel}$0.8$ & \cellcolor{ourspastel}$0.4$\\
least answer bias & \cellcolor{failpastel}$5.0$ & \cellcolor{failpastel}$13.6$ & $4.3$ & \cellcolor{failpastel}$3.8$ & $7.2$ & $6.4$\\
random & $4.3$ & $10.2$ & $5.0$ & $3.7$ & $11.1$ & $8.7$\\
\oursrule
\textbf{SRA (ours)} & \cellcolor{notepastel}$1.6$ & \cellcolor{ourspastel}$1.9$ & \cellcolor{notepastel}$2.8$ & \cellcolor{notepastel}$2.5$ & \cellcolor{ourspastel}$0.8$ & \cellcolor{ourspastel}$0.4$\\
\midrule
\multicolumn{7}{l}{\emph{reference: runs every candidate on the test set}}\\
avg.\ confidence, raw & $1.9$ & $7.1$ & $1.8$ & $0.6$ & \cellcolor{failpastel}$12.0$ & $7.0$\\
nuclear norm, raw & \cellcolor{ourspastel}$1.5$ & \cellcolor{ourspastel}$2.1$ & $1.9$ & $1.1$ & \cellcolor{failpastel}$12.0$ & $8.3$\\
best of nine estimators & \cellcolor{ourspastel}$1.5$ & \cellcolor{ourspastel}$2.1$ & \cellcolor{ourspastel}$1.5$ & \cellcolor{ourspastel}$0.5$ & \cellcolor{failpastel}$11.6$ & \cellcolor{ourspastel}$5.1$\\
best single model & $2.2$ & \cellcolor{failpastel}$16.5$ & $1.8$ & $0.5$ & \cellcolor{failpastel}$12.0$ & $7.0$\\
\bottomrule
\end{tabular*}
\end{table}

\subsection{SRA improves selection}\label{sec:selection}
\begin{figure*}[t]
\centering
\includegraphics[width=0.8\textwidth]{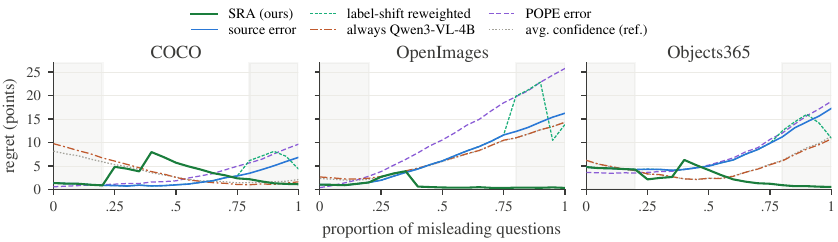}
\caption{\textbf{SRA switches to a different model once misleading questions are common, and on OpenImages and Objects365 it then stays close to the best model.} Regret of each selector on test sets resampled from each corpus's misleading test sets and calibration items at each proportion of misleading questions, $21$ candidates, with SRA given the true proportion. In the gray regions nearly all items share one label.}
\label{fig:regimeaxis}
\end{figure*}
\begin{figure*}[t]
\centering
\begin{minipage}[c]{0.68\textwidth}\centering
\includegraphics[width=\linewidth]{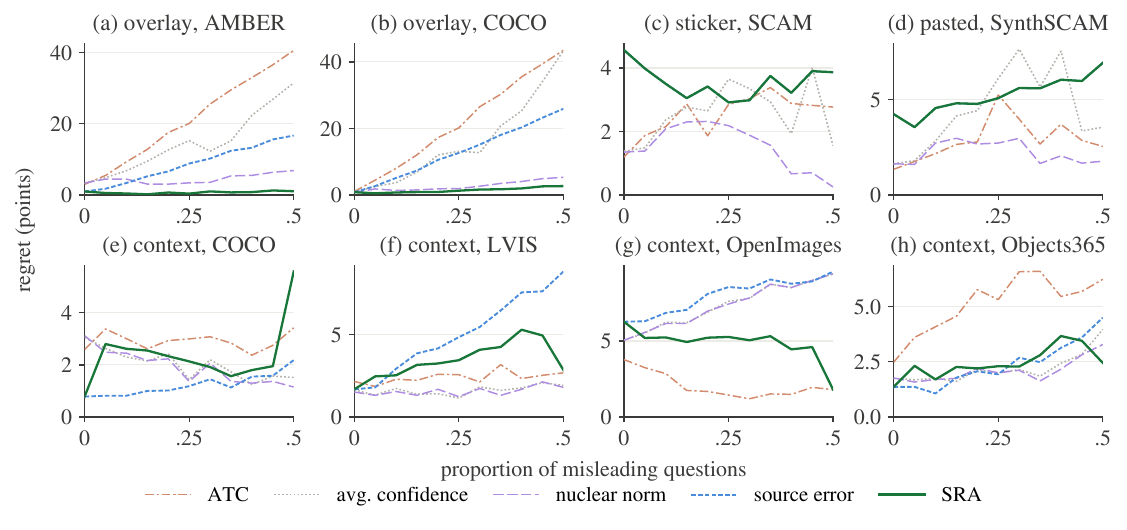}
\end{minipage}\hfill
\begin{minipage}[c]{0.30\textwidth}
\caption{\textbf{Under overlays that deny a visible object, confidence-based selectors fall far behind and SRA does not.} Mean regret on CueDial over $30$ test sets at each proportion of misleading questions, $30$ candidates and audited labels. SRA estimates that proportion with OCR in (a--d) and is given it in (e--h), whose corpora are those of \cref{fig:regimeaxis} plus LVIS. (c,\,d) Physical and pasted stickers, where the curves of SRA and source error overlap.}
\label{fig:cuedial}
\end{minipage}
\end{figure*}

Across the $87$ test sets, SRA selected a different model from source error on $29$, and on $27$ of these the selected model had lower error. Selection by source error picks InternVL3-14B on every test set. The SRA selection instead depends on the test set, picking Qwen2.5-VL-7B where most questions mislead, InternVL3-1B on most constructed test sets, and a Qwen3-VL model on some of the plausible and mixed ones (\cref{app:decisions}). Without Qwen2.5-VL-7B, its usual choice, SRA still lowered the error on $26$ of the $29$.

\begin{table}[t!]
\centering\scriptsize
\caption{\textbf{On test sets where $80\%$ of questions mislead, SRA selects close to the best model on OpenImages and Objects365 and ties source error on COCO.} $21$ candidates, error in percent, regret in points. Italic rows use target data, namely average confidence on the test set and selection by error on $400$ labeled items drawn from the test sets themselves (none where the sets hold fewer). Labeled items drawn like the test sets (\cref{tab:targetsample}) trail SRA by $1$ to $4$ points at $100$ items or fewer on OpenImages and Objects365, come within $0.4$ points at $650$, and beat it on COCO from $100$ on. On new images, only the choices of SRA and source error were fixed in advance, and SRA's reduction over source error has a $95\%$ CI above zero on OpenImages and Objects365 and spanning zero on COCO, and pooled over these two corpora's new images, error falls from $30.4\%$ to $18.8\%$.}
\label{tab:hc}
\setlength{\tabcolsep}{2pt}
\begin{tabular*}{\linewidth}{@{\extracolsep{\fill}}llcccc}
\toprule
 & & \multicolumn{2}{c}{\cref{tab:testsets} sets} & \multicolumn{2}{c}{new images}\\[-1pt]
\cmidrule(lr){3-4}\cmidrule(lr){5-6}
corpus & selector & error\,$\downarrow$ & regret\,$\downarrow$ & error\,$\downarrow$ & regret\,$\downarrow$\\
\midrule
 & source error & $6.4$ & $2.3$ & $6.6$ & $2.4$\\
 & \emph{avg.\ conf.} & \cellcolor{ourspastel}$4.8$ & \cellcolor{ourspastel}$0.6$ & \cellcolor{ourspastel}$5.3$ & \cellcolor{ourspastel}$1.2$\\
 & \emph{labeled items} & \cellcolor{ourspastel}$4.8$ & $0.7$ & \cellcolor{ourspastel}$5.3$ & \cellcolor{ourspastel}$1.2$\\
 & random & $7.8$ & $3.7$ & $7.7$ & $3.6$\\
\oursrule
\multirow{-5}{*}[0.45ex]{COCO} & \textbf{SRA (ours)} & \cellcolor{notepastel}$6.6$ & \cellcolor{notepastel}$2.5$ & \cellcolor{notepastel}$6.4$ & \cellcolor{notepastel}$2.3$\\
\midrule
 & source error & $31.2$ & \cellcolor{failpastel}$13.6$ & $31.5$ & \cellcolor{failpastel}$12.8$\\
 & \emph{avg.\ conf.} & $29.6$ & \cellcolor{failpastel}$12.0$ & $26.9$ & $8.1$\\
 & \emph{labeled items} & -- & -- & $20.0$ & $1.3$\\
 & random & $28.7$ & $11.1$ & $28.3$ & $9.5$\\
\oursrule
\multirow{-5}{*}[0.45ex]{OpenImages} & \textbf{SRA (ours)} & \cellcolor{ourspastel}$18.4$ & \cellcolor{ourspastel}$0.8$ & \cellcolor{ourspastel}$19.9$ & \cellcolor{ourspastel}$1.1$\\
\midrule
 & source error & $30.3$ & \cellcolor{failpastel}$12.5$ & $29.3$ & \cellcolor{failpastel}$12.1$\\
 & \emph{avg.\ conf.} & $24.8$ & $7.0$ & $24.6$ & $7.5$\\
 & \emph{labeled items} & $18.3$ & $0.6$ & \cellcolor{ourspastel}$17.5$ & \cellcolor{ourspastel}$0.4$\\
 & random & $26.4$ & $8.7$ & $25.4$ & $8.2$\\
\oursrule
\multirow{-5}{*}[0.45ex]{Objects365} & \textbf{SRA (ours)} & \cellcolor{ourspastel}$18.1$ & \cellcolor{ourspastel}$0.4$ & \cellcolor{notepastel}$17.6$ & \cellcolor{notepastel}$0.5$\\
\bottomrule
\end{tabular*}
\end{table}

\paragraph{Constructed, plausible, and uncurated test sets.}\label{sec:rulespope}\label{sec:audit}\label{sec:frozensel} On the constructed test sets, SRA cuts regret to the level of the best estimator that runs every candidate, though these test sets' edits resemble those of the reliance measurement, which favors SRA. On the plausible test sets it halves the regret of selection by source error (\cref{tab:calconf}). On the uncurated test sets, where misleading questions are rare, SRA keeps the model selected by source error with $21$ candidates, and with all $30$ and the shift values fixed for the held-out models, its regret there is $3.8$ points against $1.7$ for selection by source error (\cref{app:mixsrc}).

\paragraph{Misleading test sets.}\label{sec:hc}\label{sec:port} On the OpenImages and Objects365 test sets where most questions mislead, the error of SRA's choice is close to the best candidate's, whereas the choice of source error sits $13$ points above it (\cref{tab:hc}). The detector is calibrated on $650$ labeled items on these corpora and on $1{,}200$ on COCO (\cref{tab:protocol,app:procedure}). A pre-registered test on new images of both corpora confirms the selection (\cref{app:confirm}). On these two corpora, plain or reweighted accuracy on the calibration items does not beat SRA at any budget from $24$ to $400$ items (\cref{app:budget}). Mandoline ranks the models within a test set better than SRA but places a worse model first on these two corpora (\cref{app:decisions}). The fixed POPE choice matches SRA here but has mean regret $4.4$ points against $2.2$ for SRA over the $78$ test sets outside POPE and SCAM, and, post hoc, rules that switch at a single threshold on the estimated shift each fall behind SRA on some test-set type (\cref{app:decisions}). On COCO, whose misleading questions nearly every candidate answers correctly, SRA ties source error on the test sets of \cref{tab:testsets} and, against its pre-registered prediction, on new images (\cref{app:confirm,app:abserror}). Label-shift reweighting with the true yes/no balance does worse than SRA on all three corpora (\cref{tab:calconf}), so label balance does not explain the gain (\cref{fig:regimeaxis}).

\begin{figure}[t]
\centering
\includegraphics[width=\linewidth]{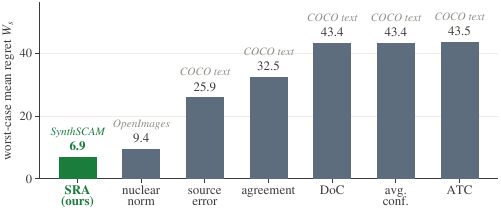}
\caption{\textbf{Without running any candidate on the target, SRA matches the nuclear norm and leads the rest in worst-case mean regret.} Each selector's largest $W_s$ over the eight CueDial families (named above each bar), public split, $30$ candidates, SRA given the true proportion on the context families. Post hoc item-level bootstrap intervals separate SRA from every selector but the nuclear norm here, and on the hidden split the nuclear norm's mean $W_s$ over families is lower by $1.7$ points ($[0.6,2.9]$, \cref{app:cuedial}).}
\label{fig:robust}
\end{figure}

\subsection{Selection on CueDial}\label{sec:dialresults}
No selector stays close to the best model on every family (\cref{fig:cuedial}). When overlaid text denies a visible object, the models it misleads remain confident (mean confidence $.86$, against $.94$ on clean items). Confidence-based selectors such as ATC therefore reach a worst-case mean regret $W_s$ above $40$ points (\cref{fig:robust}), and selection by source error trails the best model by $17$ to $26$ points (\cref{fig:cuedial}a,b, \cref{tab:cuedialhidden}). SRA, which estimates the proportion with OCR, keeps $W_s$ below $3$ points on the two overlay families. On SCAM's stickers, SRA and source error miss the best model, and the nuclear norm does best.

Under scene context, where SRA is given the true proportion, estimators that run every candidate do best on LVIS, OpenImages, and Objects365, where SRA improves on source error but trails them, and on COCO and the OpenImages hidden split SRA trails source error (\cref{fig:cuedial}e--h and \cref{tab:cuedialhidden}).

\section{Conclusion and Limitations}\label{sec:conclusion}
Reliance, measured once on edited source images against a matched control, predicts which VLMs a cue will mislead on existing benchmarks. Where SRA selected a different model from source error, that model had lower error on $27$ of $29$ test sets, and pre-registered tests on new images confirmed the gain on Objects365 and OpenImages but not on COCO, where SRA tied source error. When overlaid text may deny an object, confidence-based selectors can fail badly, and CueDial's cached answers and hidden split let future selectors be compared without running a model.

\paragraph{Limitations.} On COCO, where most misleading questions cause few errors, SRA switches models too early (\cref{fig:regimeaxis}), and its score, read as an error, is several times too high there and too high elsewhere once labels are audited (\cref{app:abserror}). On SCAM, source error does not carry over, so SRA misses the best model (\cref{app:scam}). Cue effects are assumed additive. The correction assumes each question type triggers the detector as often on the target as on the calibration set. Labeled items drawn like a test set beat SRA on COCO and come within $0.4$ points of it on OpenImages and Objects365 at $650$ items (\cref{tab:targetsample}), but there one set of $650$ labeled items serves SRA on all of a corpus's test sets, ordinary or misleading (\cref{app:confirm}). We test object presence, not attributes or counts.

{\small\bibliographystyle{ieeenat_fullname}\bibliography{SRA_v45}}

\clearpage
\clearpage
\appendix
\section*{Overview of the appendix}
\Cref{app:analysis} gives the mixture model behind the score and why a ranking must depend on the target. \Cref{app:wbdetail} reports the Waterbirds validation behind \cref{fig:formulation}, and \cref{app:cocodetail} the prediction ablations, the estimators that run every candidate on the test images, and the sticker tests of \cref{sec:transfer}. \Cref{app:reversaldetail} describes the reversal test on the constructed test sets and the overlay templates. \Cref{app:auditfrozen} holds the selection experiments of \cref{sec:selection,sec:port}, from the test sets of \cref{tab:testsets} and the two new corpora to the tests on new images, the label budget, the estimator baselines, and the score's level against realized error. \Cref{app:dayzero} reports the nine held-out models, \cref{app:cuedial} the construction and audit of CueDial, and \cref{app:repro} the implementation. The pre-registration records, hashes, and outcomes are in the project repository~\todo{URL}.

\section{Theory}\label{app:analysis}

\subsection{The mixture model behind the score}\label{app:theory}
Fix a model $f$ and one cue $j$. A target item is agreeing when the cue agrees with its answer and misleading when the cue points to the wrong answer, and $d_j$ is the proportion of misleading items. If items of each kind are exchangeable between source and target, the model's target error is exactly linear in $d_j$,
\begin{equation}
\err_f(d_j)=(1-d_j)\,\err_f^{=}+d_j\,\err_f^{\neq}=\err_f^{=}+d_j\,r_{f,j},
\label{eq:law}
\end{equation}
with $\err_f^{=}$ the error on agreeing items, $\err_f^{\neq}$ the error on misleading ones, and $r_{f,j}=\err_f^{\neq}-\err_f^{=}$ the extra error on misleading questions. \Cref{eq:sra} estimates $\err_f^{=}$ with source error, measured on source questions on which no cue misleads, and $r_{f,j}$ with reliance $R_{f,j}$ (\cref{eq:flip}). An edit turns a source item into a misleading question with the same content, and the matched control removes what the edit itself changes. With several cues, \cref{eq:sra} adds one term per cue, which assumes that the cues do not interact. \Cref{app:wbdetail} measures how far compound shifts fall below this sum. The quantities involved are the following.

\begin{center}\footnotesize
\setlength{\tabcolsep}{4pt}
\begin{tabular}{@{}l>{\raggedright\arraybackslash}p{3.2cm}>{\raggedright\arraybackslash}p{2.9cm}@{}}
\toprule
quantity & meaning & source\\
\midrule
$\rr_f$ & extra error when cues mislead & needs target labels\\
$\mathbf R_f$ & measured reliance & source edits with controls\\
$\dd$ & true proportions of misleading items & needs target labels\\
$\hat\dd$ & estimated shift & unlabeled target images\\
$\err_f(\mathbf 0)$ & source error & labeled source data\\
$\langle\mathbf R_f,\hat\dd\rangle$ & reliance term & $\mathbf R_f$ and $\hat\dd$ above\\
\bottomrule
\end{tabular}
\end{center}

\begin{remark}[What the score does and does not give]
\label{rem:gap}
\Cref{eq:law} is exact for the latent pair $(\rr_f,\dd)$, and
\cref{eq:sra} substitutes the measured pair
$(\mathbf R_f,\hat\dd)$. Fixing the model and writing
$\mathbf R,\rr$ for its two vectors, the gap
$\Delta:=\langle\mathbf R,\hat\dd\rangle-\langle\rr,\dd\rangle$
decomposes, and is bounded, as
\begin{align*}
\Delta&=\langle\mathbf R,\hat\dd-\dd\rangle
+\langle\mathbf R-\rr,\dd\rangle,\\
|\Delta|&\le\|\mathbf R\|_\infty\,\|\hat\dd-\dd\|_1
+\|\mathbf R-\rr\|_\infty\,\|\dd\|_1.
\end{align*}
The first term is detector error, which we measure wherever the
composition is known (\cref{app:confirm,app:rsrc,app:organic}). The second is
measurement error in reliance. It combines sampling error (\cref{app:noise}) with dependence on the overlay template, which the spread across templates bounds. Three further deviations come from the intercept, from the additivity assumption, and from edit artifacts. Source error is measured on a balanced set, so on a target whose present proportion is $\pi$ the prediction is off by $\tfrac12-\pi$ times the gap between the model's error on present objects and its false-yes rate on absent ones, a constant per model that an intercept reweighted to the target's balance removes (\cref{app:regretfrac} tests that variant). Cues interact (\cref{app:wbdetail}), and edit artifacts are bounded by the matched controls, placebo edits, and the check for traces of the erased object (\cref{app:artifacts}). We measure or bound each term separately and claim no combined bound, so the mixture model motivates \cref{eq:sra} and the experiments validate it as a ranking score.
\end{remark}

\subsection{Why a ranking must depend on the target}\label{app:scalar}\label{app:proofs}
A score that gives each model one number, whatever the target domain, orders the models the same way in every target. Suppose two models swap order between two targets, with $\err_a(\dd)>\err_b(\dd)$ but $\err_a(\dd')<\err_b(\dd')$. Any such score then ranks the pair wrongly in one of the two targets, so the number of pairs that swap between targets is a lower bound on the ranking errors of every score that ignores the target. \Cref{eq:sra} avoids this bound only through its reliance term.

\section{Waterbirds Validation Detail}\label{app:wbdetail}\label{app:mainfloats}\label{app:waterbirds}
On Waterbirds the quantity predicted is the accuracy drop, so the prediction is the reliance term $\langle\mathbf R_f,\hat\dd\rangle$ alone, without source error. \Cref{fig:formulation} illustrates the full score of \cref{eq:sra} on two of the ten Waterbirds models (\cref{app:impl}). On its target domain, each unit of shift adds $.31$ error to Qwen2.5-VL-3B and $.04$ to LLaVA-1.5-7B, whose realized errors are $.538$ and $.488$. Over the informative graded and constructed targets, the reliance term predicts each model's drop and selects the model with the smallest drop better than clean accuracy, which predicts the drop as its negative (\cref{tab:select}). Graded targets apply each cue to a random proportion of the items, one cue at proportions $.25$, $.5$, $.75$, and $1$, two or three cues at $.5$ each, or four uneven mixes, and constructed targets apply a cue or a combination of cues to every item. Our row uses the detectors' estimated shift on the graded targets and the constructed shift on the constructed targets. With the constructed shift, graded $\tau$ is $+.72$. Background swaps barely change accuracy, so the background coordinate carries little signal. On the $25$ constructed compound cells, model and target pairs, the compound drop differs from the sum of the single-cue drops by $.14$ on average (median $.08$). In $36\%$ of pairs it falls more than $.1$ below that sum and in $8\%$ above it, as expected when drops approach their ceiling.

\begin{table}[t]
\centering\small
\caption{\textbf{Waterbirds drop prediction and selection.}
The graded columns give the regret, in drop units, of the model with the smallest predicted drop and the pooled Kendall $\tau$ between predicted and realized drops, over the informative graded targets (those whose realized drops have a standard deviation across models of at least $.10$, the text and prompt targets at proportions $.75$ and $1$). The constructed columns give the regret and the mean per-target $\tau$ over the four informative constructed targets (text, background and text, prompt, background and prompt). Dashes mark selectors we did not evaluate in a setting. Our row is the reliance term alone. Green marks the best value per column, and yellow the reliance term where it is not the best.}
\label{tab:select}
\setlength{\tabcolsep}{4pt}
\begin{tabular}{lcccc}
\toprule
& \multicolumn{2}{c}{graded} & \multicolumn{2}{c}{constructed}\\
\cmidrule(lr){2-3}\cmidrule(lr){4-5}
selector & regret $\downarrow$ & $\tau$ $\uparrow$ & regret $\downarrow$ & $\tau$ $\uparrow$\\
\midrule
clean accuracy & $.020$ & $+.07$ & $.027$ & $+.31$\\
mean reliance & -- & -- & $.027$ & $+.45$\\
random & $.215$ & -- & $.234$ & --\\
\oursrule
\textbf{reliance term only (ours)} & \cellcolor{ourspastel}$.016$ & \cellcolor{ourspastel}$+.62$ & \cellcolor{ourspastel}$.023$ & \cellcolor{ourspastel}$+.73$\\
\bottomrule
\end{tabular}
\end{table}

\section{Prediction Detail}\label{app:cocodetail}

\subsection{Prediction ablations}\label{app:ablate}
\Cref{tab:ablate} shows that the correlation of \cref{sec:transfer} survives each ablation. We claim it only for reliance averaged over objects, because it is weaker for individual objects.

\begin{table}[ht]
\centering\small
\caption{\textbf{Ablations of the prediction result} ($18$ VLMs,
reliance measured by edits against hallucination on unedited images). The bootstrap
$95\%$ CI for the Pearson correlation with answer bias subtracted is $[.77,.96]$, and
resampling clustered by model family gives $[.78,.96]$. The leave-one-family-out row gives the range over families. Residuals are those of each variable after regressing on answer bias. The leave-one-family-out and within-family rows use the raw outcome, and families group the two OneVision models with Qwen, their language backbone. Green marks the column maximum, and yellow the chosen variant where it is not.}
\label{tab:ablate}
\begin{tabular}{lcc}
\toprule
variant & Pearson $\uparrow$ & Spearman $\uparrow$\\
\midrule
raw & $+.75$ & $+.78$\\
\textbf{answer bias subtracted} & \cellcolor{ourspastel}$+.89$ & \cellcolor{ourspastel}$+.86$\\
residuals given bias & $+.88$ & $+.81$\\
leave-one-family-out & $+.69$--$.77$ & $+.73$--$.79$\\
within-family only & $+.62$ & $+.78$\\
\bottomrule
\end{tabular}
\end{table}

\subsection{Predicting hallucination against estimators that run on the target}\label{app:pred}
The reference entries of \cref{tab:pred} use the answers already collected for \cref{fig:transferplane,fig:dayzero}. On the COCO hallucination set, each score uses each model's answers on the $558$ questions and predicts the raw hallucination rate. On DASH-B it uses the $1{,}341$ images without the queried object and predicts the false ``yes'' rate, and on SCAM the two SCAM test sets and the false ``yes'' rate on the sticker word minus the same rate on the clean copies (\cref{app:scam}). ATC sets its threshold at the confidence quantile that reproduces each model's error on its $300$ labeled source items ($100$ present objects, $100$ erased, $100$ neutral absent), and DoC is the error on those items plus the drop in mean confidence from them to the target. Average confidence, entropy, the nuclear norm, and MaNo follow \cref{app:lfall}, signed so that a larger value predicts more hallucination. The partial row ranks reliance, failure, and answer bias and correlates the residuals of the first two after regressing out the third. For the nine held-out models, bootstrap $95\%$ CIs of reliance's Spearman correlation are $[.16,1.00]$ on DASH-B (one-sided permutation $p{=}.008$) and $[.58,1.00]$ on SCAM ($p{<}.001$). Reliance computed from hard yes/no answers instead of probabilities ranks the $30$ models almost identically (Spearman $.96$), so models that return only text can be measured.

\subsection{A second cue type: physical stickers}\label{app:scam}
SCAM~\cite{westerhoff2025scam} holds $1{,}162$ photographs of objects ($660$ labels) carrying a physical sticker or sign with a misleading word ($206$ words). Each photograph also has a clean copy (NoSCAM) and a copy with the word pasted digitally (SynthSCAM). We asked every model two questions per image, ``is there a [sticker word]?'' (no) and ``is there a [object label]?'' (yes). We excluded two images whose sticker word is contained in their object label. Splitting SCAM, NoSCAM, and SynthSCAM in two by image id then gives six test sets of $1{,}158$ to $1{,}162$ questions, $6{,}960$ per model. The composition is the dataset's. The measurement, the detector, and the predictions for all $30$ candidates were fixed before any candidate answered a SCAM question.

\paragraph{Measurement.} On each of the $100$ source images, we paste a sticker-like label that names an absent object and ask about that object. Reliance is the yes rate under that label minus the yes rate under an agreeing label that names the present object, the matched control of \cref{eq:flip}, averaged over two label styles.

\paragraph{Detector.} EasyOCR flags a question when its word appears in the recognized text, as the union of two OCR passes with fuzzy matching. It flags $.80$ of questions about the sticker word on SCAM and at most $.005$ of the others, and gives an uncorrected flagged proportion $\hat d_{\mathrm{text}}$ of $.40$ on SCAM, $.50$ on SynthSCAM, and $.00$ on NoSCAM.

\paragraph{Score.} $\widehat{\err}_f=\err_f(\mathbf 0)+R_f\,\hat d_{\mathrm{text}}$, with source error over the $100$ questions about present objects. With the source error of \cref{eq:sra}, SRA selects the same models as source error on all six SCAM test sets.

Reliance measured with pasted labels predicts the stickers' effect (\cref{tab:pred}), but selection on SCAM does not improve on source error, because SRA adds the right reliance term to a source error that does not carry over. Source error over present objects has Spearman $-.04$ with error on the clean copies and $-.01$ with error under the stickers. SRA therefore keeps InternVL3-14B (SCAM error $.163$), while the best model is Qwen3-VL-8B ($.115$), which COT and every confidence-based estimator but DoC select (DoC selects LensVLM-9B). Text reliance measured with overlays that deny the object, averaged over the four templates, has the wrong sign on SCAM ($-.50$). Asserting a sticker word and negating an object are different cues, and only the matching one transfers.

\subsection{Free-text naming under physical stickers}\label{app:scamopen}
We asked $40$ models to name the object in SCAM photographs, ``What object is this? Answer in one or two words.'' (greedy, at most $8$ new tokens). They are the $30$ candidates and ten other models, InternVL3.5-2B, -4B, and -8B~\cite{wang2025internvl35}, LFM2-VL-1.6B and -3B~\cite{lfm2vl2025}, Granite-Vision-3.3-2B~\cite{granitevision2025}, Pixtral-12B~\cite{agrawal2024pixtral}, SmolVLM2-500M~\cite{marafioti2025smolvlm}, GLM-4.1V-9B-Base~\cite{glm41v2025}, and LLaVA-Interleave-Qwen-7B~\cite{li2024llavainterleave}. We used the $1{,}160$ SCAM images of \cref{app:scam} and split them by a fixed seed into $200$ source and $960$ target images. An answer is correct if it equals the object label after normalization or if a judge, Qwen2.5-14B-Instruct~\cite{qwen25report} in 4-bit, says it names the labeled object. On $300$ randomly drawn scored answers the judge agreed with a reading of the label and answer on $95\%$. On each source image we paste the sticker word and, as the matched control, the true label, in the two label styles of \cref{app:scam}. Reliance is the error under the sticker word minus the error under the control, and source error is the error on the clean source images. The detector flags an image when EasyOCR reads a word that names an object from the label set other than CLIP's top label, with sensitivity $.635$ and false-positive rate $.035$ on the source split, and gives a corrected estimate (\cref{sec:dhat}) $\hat d$ of $.97$ on SCAM and $.01$ on NoSCAM. The predictions and each selector's choice for every target test set were fixed before any model answered a target image.

Reliance measured with pasted words predicts the damage of the physical stickers, Spearman $.79$ over the ten models outside the candidates and $.74$ over all $40$, and the SRA score ranks the models on SCAM better than clean accuracy ($.85$ against $.70$). Selection is again limited by source error. The target test sets mix the sticker and clean copies of the $960$ target images at five sticker rates. The largest mean regret of SRA over the five rates is $.084$ against $.077$ for clean accuracy, since at the mixed rates SRA chooses LFM2.5-VL-3B, whose clean error on the $200$ source images ($.235$) is $5$ points below its error on the $960$ target images ($.286$), while the reliance term orders it correctly against clean accuracy's choice. Reliance measured on COCO yes/no questions does not order the models' free-text reliance (Spearman $-.02$) and moves the prediction by $.04$ to $.14$, so both terms must be measured in the answer format of the target.

\section{Overlay Templates and Predicted Reversals}\label{app:reversaldetail}\label{app:crossfam}\label{app:constructed}
On the ten constructed test sets, predictions from source edits named the model pairs that should swap order as the mix of misleading cues changes, and the pairs swapped far more often than under shuffled predictions on both AMBER and constructed COCO (\cref{tab:frozen}), precision being the fraction of predicted swaps that occurred.

\paragraph{Test sets.} The protocol uses five test sets of $200$ AMBER~\cite{wang2023amber} yes/no questions, and none of their images was used in any measurement, calibration set, or test set outside CueDial. Each item is one of four types. The object is present and untouched (answer yes), present with a misleading overlay (yes), absent but listed by AMBER as a likely hallucination for an image with at least three labeled objects (no), or absent and unrelated to the scene (no). Every test set has $100$ yes items and $100$ no items, and only the split of the $100$ misleading items between overlay and context varies. The five constructed COCO test sets follow the same design, except that a context item asks about an absent object with at least two of its partners present. The test overlay's wording and position differ from all four measurement templates, and it shares its colors with one of them and its font with another.

\begin{table}[ht]
\centering\footnotesize
\caption{\textbf{Reversal test outcomes} on the AMBER test sets (left of each slash) and the constructed COCO test sets (right). Interaction $r$ is the correlation with observed error after removing model and test-set means, using the shift estimated from unlabeled images. Precision uses the known composition. $95\%$ bootstrap CIs over models for precision on AMBER are $[.21,.58]$ and $[.32,.72]$ (rows in order), and for interaction on constructed COCO $[+.03,+.79]$ and $[+.49,+.89]$. The permutation nulls for precision are $.16$ and $.14$ ($95$th percentile $.22$ on AMBER). Green marks the best value in a column.}
\label{tab:frozen}
\begin{tabular}{lcc}
\toprule
reliance estimate & interaction $r$ $\uparrow$ & precision $\uparrow$\\
\midrule
single template & $.37\,/\,.53$ & $.41\,/\,.33$\\
\oursrule
\textbf{averaged over four templates} & \cellcolor{ourspastel}$.64\,/\,.76$ & \cellcolor{ourspastel}$.52\,/\,.46$\\
\bottomrule
\end{tabular}
\end{table}

\begin{figure}[ht]
\centering
\includegraphics[width=0.9\linewidth]{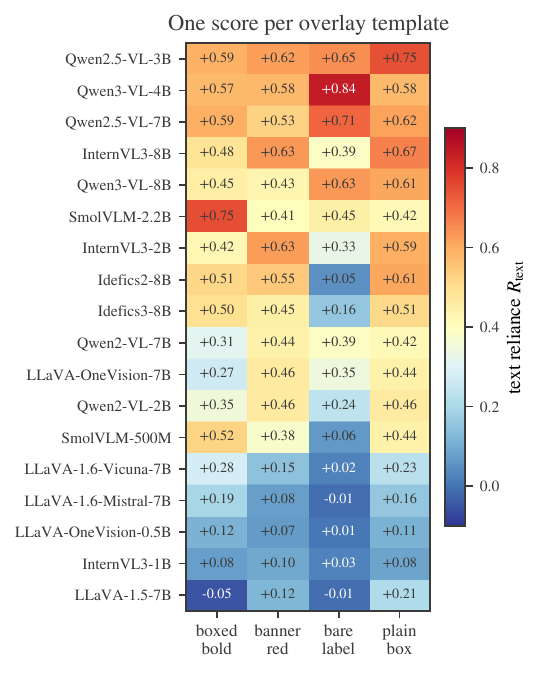}
\caption{\textbf{Text reliance depends on how the text is drawn.}
Each column is one overlay template (boxed bold ``no obj here'',
red banner sentence, bare label ``obj: absent'', plain box sentence),
and each row is one model, sorted by mean. The bare label is the weakest template for most models, and full sentences mislead
many more. The Qwen family responds to every template, the bare label included.}
\label{fig:opmatrix}
\end{figure}

\paragraph{Overlay templates.} \Cref{fig:opmatrix} shows why we average text reliance over four templates. A single template can understate a model badly. For six models, the held-out test template changes the answer $1.6$ to $2.3$ times as often as a single template predicts.

\section{Selection Detail}\label{app:auditfrozen}
\begin{figure}[ht]
\centering
\includegraphics[width=\linewidth]{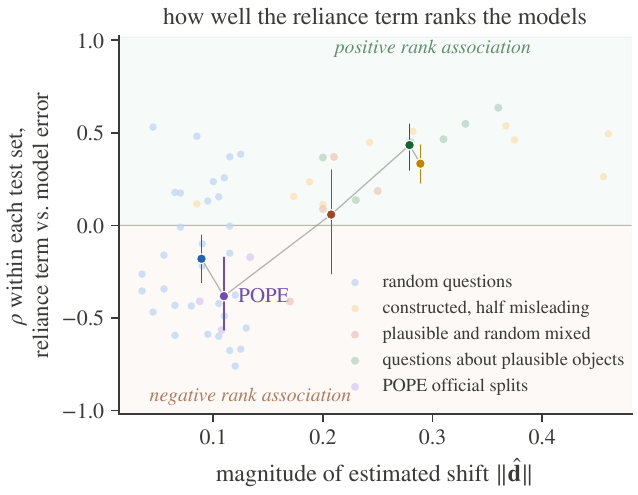}
\caption{\textbf{The reliance term orders the models once misleading questions are common.} Spearman correlation within each test set of the reliance term with realized error against the estimated shift (type means, $95\%$ whiskers). Above a shift of about $0.2$ it orders the models, and source error orders them on POPE.}
\label{fig:regimea}
\end{figure}

The plausible and misleading test sets ask about an object only when at least two of its partners (\cref{sec:discovered}) are present, as a checklist would. On the uncurated test sets few questions mislead. Misleading context questions make up $2.2\%$ of items by annotation count, and image text appears on $32\%$ of items but never names and negates the queried object. \Cref{fig:regimea} shows how the reliance term's correlation with error within each test set rises with the estimated shift.

\subsection{Noise and bias in the shift estimate}\label{app:noise}
Gaussian noise of $\sigma$ up to $.2$ on the shift coordinates, which lie in $[0,1]$, raises mean selection regret on the Waterbirds graded targets from $.016$ to at most $.020$. Systematic detector bias is the harder case. We tested it on the $42$ uncurated and constructed test sets with the $18$ development models by perturbing each estimated shift $\hat\dd$ and recomputing the SRA score. Multiplicative miscalibration ($\times0.5$, $\times1.5$) leaves mean regret at $.013$--$.016$, and so does inflating only the context coordinate by $+0.10$, the direction of the context detector's false-positive bias. A uniform bias of $+0.10$ on every coordinate doubles regret to $.028$, still better than source error alone ($.043$) or the reliance term alone ($.069$). Where the shift is small, a biased reliance term must first overcome the gap in source error before it changes the selection.

\paragraph{Target sample size.} Re-estimating $\hat\dd$ from
$n$-item subsamples of each test set leaves selection nearly unchanged, again with the $18$ development models.
We drew $500$ hypergeometric samples of the detector's flag counts per test set and size, and we always scored regret on the full test set. Regret on the constructed test sets stays at $.018$--$.019$ from $n{=}200$ down to $n{=}20$, where the selection changes on $3.8\%$ of draws, and on the uncurated test sets it stays at $.012$ for every $n$.

\paragraph{Reliance sample size.} Each model's context reliance comes
from $100$ source items. Standard errors from an item bootstrap ($B{=}4{,}000$) are about half the between-model standard deviation for context reliance, with $.033$ at the median model and $.042$ at most against a standard deviation of $.073$ across the $21$ models. For text reliance they are about a seventh of it ($.030$, at most $.046$, against $.21$). The pair that decides the selection when most questions mislead, InternVL3-14B and Qwen2.5-VL-7B, differs by $.21$ in context reliance ($95\%$ CI $[.12,.31]$).

\subsection{Protocol on the two new corpora}\label{app:newcorpora}\label{app:oiarm}\label{app:o365arm}
On OpenImages and Objects365 the detector combines two signals per item, an object detector's absence call and CLIP's score for the object's usual context (\cref{app:procedure}). Detectors built on a contrastive encoder alone score plausibly absent objects like present ones, a blind spot they share with the candidates. On the calibration set and the random absent items we measure, for each item type (misleading, present, random absent), how often each combination of the two signals occurs, and nonnegative least squares then estimates each test set's mix of the three item types from these rates, as in mixture proportion estimation. SRA beats selection by source error on both corpora and ranks the models within their test sets (\cref{tab:hc}, \cref{app:decisions}), which a source error over present objects alone does not (\cref{app:mixsrc}). Measuring source error and reliance directly on each corpus's labeled calibration set also improves the ranking, as the OpenImages paragraph below shows.

\begin{table}[ht]
\centering\footnotesize
\caption{\textbf{Protocol on the two new corpora.} For Objects365 we fixed one shard of the train split, chosen from annotation counts alone. The random absent items give the detector's signal rates on that item type.}
\label{tab:protocol}
\setlength{\tabcolsep}{2pt}\scriptsize
\begin{tabular}{@{}l>{\raggedright\arraybackslash}p{2.95cm}>{\raggedright\arraybackslash}p{2.75cm}@{}}
\toprule
& OpenImages & Objects365\\
\midrule
split & validation & one train shard\\
query vocabulary & $25$ categories a COCO object detector detects & $29$ mapped categories\\
per category & $\ge50$ verified present, $\ge50$ verified absent & $\ge40$ misleading, $\ge40$ present\\
partners mined over & $423$ categories & $365$ categories\\
calibration set & $400$ ($150$ misleading, $250$ present) & $400$\\
random absent items & $250$ & $250$\\
misleading test sets & $2\times125$, fraction $.80$ & $4\times200$, fraction $.80$\\
ordinary test sets & $3\times120$ & $3$ (proportions $0$, $0$, $.01$)\\
detector & object detector $+$ partner context (\cref{app:procedure}) & same, rates measured on its calibration set and random absent items\\
\bottomrule
\end{tabular}
\end{table}

\paragraph{OpenImages.} The OpenImages test sets use the validation split, whose image-level labels are human-verified in both directions, so absence is a labeled fact. A misleading item is a verified absent object with at least two of its top-five partners verified present. We drew the calibration set, the random absent items, and the test sets of \cref{tab:protocol} from disjoint images, with the item lists and detector constants fixed before any candidate ran. The selections reported for these test sets use the source error of \cref{eq:sra}, adopted after their answers were seen (\cref{app:dayzero}), and are retrospective. \Cref{app:confirm} gives the prospective test. With a source error over present objects alone, the mean Spearman correlation within each test set, $\bar\rho$, is $-.08$ on OpenImages and $-.05$ on Objects365 under the true composition, against $+.46$ and $+.60$ with the source error of \cref{eq:sra} and the estimated shift (\cref{app:decisions}), and measuring source error and reliance on the calibration set gives $+.78$ and $+.86$.

\paragraph{Objects365.} The shard, vocabulary, partners, calibration set, and test sets of \cref{tab:protocol}, and the absence convention of COCO train2017, under which an object is absent when it is not annotated, were fixed before any image was downloaded, any detector scored, or any candidate run. The detector is the OpenImages detector with the rates for each item type measured on the Objects365 calibration set and random absent items. SRA selects Qwen2.5-VL-7B on the misleading test sets (\cref{sec:hc}, \cref{tab:decisions}). Plain accuracy on the calibration set ties two models and, with the tie broken by name, has regret $.086$ against $.004$ for SRA (\cref{fig:budget} scores ties as the mean of the tied models), while reweighted accuracy selects Qwen2.5-VL-3B and does better than plain accuracy, though not than SRA (\cref{fig:budget}). On the ordinary test sets, SRA keeps the model selected by source error, while reweighted and plain accuracy on the calibration set do not (regret $.017$ and $.020$ against $.012$).

\subsection{Test on new images}\label{app:confirm}
To test the score of \cref{eq:sra} on images that no other experiment uses, we fixed the score, the candidates, the item rules, the detector, and the predictions before any image was downloaded. (The held-out predictions of \cref{app:dayzero}, made earlier, used a source error over present objects alone.) The candidates are the $21$ selection candidates (primary) and all $30$ (secondary). The two corpora are a second Objects365 train shard and the test split of OpenImages V7. Query vocabularies ($28$ Objects365 categories qualify in the second shard, against $29$ in \cref{tab:protocol}, and the $25$ OpenImages categories), co-occurrence partners, and item types follow \cref{tab:protocol}. Each corpus has four misleading test sets of $200$ items, three ordinary test sets, a $400$-item calibration set, and $250$ random absent items, all on disjoint images, read by the detector of \cref{app:newcorpora} with the rates for each item type measured on each corpus's calibration and random absent items. The test-set item lists were fixed before the detector ran, and the detector's estimates and every selection before any candidate answered a test item.

\Cref{tab:confirm} gives the outcome. SRA chose Qwen2.5-VL-7B on all eight misleading test sets, where source error chose InternVL3-14B, and Qwen2.5-VL-7B was the best model or within $3$ points of it on each. On the ordinary test sets, SRA kept the model selected by source error, which trails the best model by $.029$ and $.061$. Pooled over both corpora, the error of the selected model fell from $.304$ to $.188$ ($95\%$ CI of the reduction $[.097,.136]$, and $[.090,.144]$ on OpenImages and $[.088,.145]$ on Objects365 alone), and to $.204$ with all $30$ candidates (CI $[.081,.118]$).

\begin{table}[ht]
\centering\scriptsize
\caption{\textbf{Test on new images}, $21$ candidates. The misleading test sets A--D and the mean over the three ordinary ones, with the detector's estimated proportion of misleading questions $\hat d$ (true proportion $.80$ on A--D), the error of the best model and of the choices of SRA and source error, and the Spearman correlation $\rho$ between the SRA score and the realized error. Predictions of the reduction of error over source error, pooled and within each corpus, of a mean $\rho$ above zero within the misleading sets, and of the choices on the ordinary sets were fixed in advance and held, also with all $30$ candidates (paired bootstrap over items, $B{=}10{,}000$).}
\label{tab:confirm}
\setlength{\tabcolsep}{3pt}
\begin{tabular*}{\linewidth}{@{\extracolsep{\fill}}lccccc}
\toprule
test set & $\hat d$ & best & SRA & source & $\rho$\\
\midrule
Objects365 A & $.65$ & $.175$ & $.180$ & $.310$ & $+.75$\\
Objects365 B & $.65$ & $.170$ & $.180$ & $.285$ & $+.52$\\
Objects365 C & $.63$ & $.175$ & $.180$ & $.265$ & $+.68$\\
Objects365 D & $.41$ & $.165$ & $.165$ & $.310$ & $+.17$\\
OpenImages A & $.82$ & $.175$ & $.205$ & $.320$ & $+.52$\\
OpenImages B & $.67$ & $.205$ & $.205$ & $.340$ & $+.63$\\
OpenImages C & $.71$ & $.205$ & $.220$ & $.305$ & $+.55$\\
OpenImages D & $.72$ & $.165$ & $.165$ & $.295$ & $+.40$\\
\midrule
Objects365 ordinary & ${\le}.07$ & $.023$ & $.052$ & $.052$ & $-.30$\\
OpenImages ordinary & $.00$ & $.064$ & $.125$ & $.125$ & $-.33$\\
\bottomrule
\end{tabular*}
\end{table}

\paragraph{COCO.} We ran the same test on COCO, where SRA and source error tie on the test sets of \cref{tab:testsets}, again with everything fixed before any image was selected. It drew six test sets of $200$ items ($160$ misleading, $40$ present) from train2017 images that no other experiment uses and read them with the COCO detector of \cref{sec:dhat}, its constants unchanged. It drew no ordinary test sets, because that detector reads absence alone and holds only for test sets sampled like its calibration set. SRA chose Qwen2.5-VL-7B on all six test sets and source error chose InternVL3-14B. Their errors are statistically tied, with $21$ and with all $30$ candidates (error $.064$ for SRA against $.066$ for source error, CI of the reduction $[-.015,.018]$), so neither the predicted reduction nor non-inferiority within one point was established. The SRA score ranks the models within each test set, as predicted (mean $\rho$ of $+.46$, \cref{tab:confirmcoco}). A post hoc analysis in \cref{app:abserror} traces the tie to COCO's misleading questions, which nearly every candidate answers correctly.

\begin{table}[ht]
\centering\scriptsize
\caption{\textbf{Test on new COCO images}, $21$ candidates, columns as in \cref{tab:confirm}.}
\label{tab:confirmcoco}
\setlength{\tabcolsep}{3pt}
\begin{tabular*}{\linewidth}{@{\extracolsep{\fill}}lccccc}
\toprule
test set & $\hat d$ & best & SRA & source & $\rho$\\
\midrule
COCO A & $1.00$ & $.045$ & $.080$ & $.050$ & $+.17$\\
COCO B & $.84$ & $.025$ & $.035$ & $.050$ & $+.58$\\
COCO C & $1.00$ & $.040$ & $.065$ & $.065$ & $+.58$\\
COCO D & $1.00$ & $.045$ & $.055$ & $.070$ & $+.54$\\
COCO E & $.84$ & $.045$ & $.075$ & $.085$ & $+.58$\\
COCO F & $.59$ & $.050$ & $.075$ & $.075$ & $+.33$\\
\bottomrule
\end{tabular*}
\end{table}

\subsection{Decisions, candidate pools, and a fixed POPE choice}\label{app:decisions}
These analyses use the cached answers of the $81$ test sets outside SCAM and the source error of \cref{eq:sra}. On the test sets of \cref{tab:testsets}, Mandoline ranks the models within a test set better than SRA, with mean Spearman correlations of $.55$, $.82$, and $.83$ against $.51$, $.46$, and $.60$ on COCO, OpenImages, and Objects365, but places a worse model first on OpenImages and Objects365, with regret $3.2$ and $3.6$ points against $0.8$ and $0.4$ for SRA, and ties SRA on COCO ($2.4$ against $2.5$). The $29$ changed selections come from four model pairs (\cref{tab:decisions}). All move away from the model selected by source error, InternVL3-14B, to Qwen2.5-VL-7B on the misleading test sets and the four Objects365 plausible sets, to InternVL3-1B on eight of the ten constructed test sets, and to Qwen3-VL-2B on plausible, mixed, DASH-B, and VQAv2 test sets. Over $1{,}000$ random $15$-model subsets of the $21$ candidates, SRA changed a median of $29$ selections, lowered the error on a median of $93\%$ of them, and lowered the mean error in every subset. Without Qwen2.5-VL-7B, SRA switches to Qwen3-VL-4B on those test sets and still lowers the error on $26$ of its $29$ changes. Always choosing the model with the lowest false ``yes'' rate on POPE's adversarial split selects Qwen2.5-VL-7B, so it matches SRA on the misleading OpenImages and Objects365 test sets and does better on DASH-B and the ordinary test sets, but it falls behind source error on the uncurated and constructed test sets (\cref{tab:calconf}). Pooled over the $78$ test sets outside POPE and SCAM, where that choice is not in sample, its mean regret is $4.4$ points against $2.2$ for SRA and $4.6$ for source error. The bootstrap CI of its difference from SRA over test sets is $[1.0,3.6]$ points, and SRA is better on $41$ test sets and the fixed choice on $17$.

\begin{table}[ht]
\centering\scriptsize
\caption{\textbf{Model pairs behind the selections SRA changed}, $21$ candidates. Mean error reduction in points over the test sets of each pair.}
\label{tab:decisions}
\setlength{\tabcolsep}{2pt}
\begin{tabular*}{\linewidth}{@{\extracolsep{\fill}}llccl}
\toprule
source error & SRA & sets & lower & reduction (points)\\
\midrule
InternVL3-14B & Qwen2.5-VL-7B & $14$ & $12$ & $5.5$\\
InternVL3-14B & InternVL3-1B & $8$ & $8$ & $11.2$\\
InternVL3-14B & Qwen3-VL-2B & $6$ & $6$ & $2.1$\\
InternVL3-14B & tie of two & $1$ & $1$ & $8.0$\\
\bottomrule
\end{tabular*}
\end{table}

\paragraph{A single switch threshold.} A simpler rule keeps the choice of source error until the larger of the estimated proportions of misleading questions over the two cues passes a threshold, and then switches to the fixed POPE choice, to the candidate with the lowest reliance on that cue, or to the candidate with the lowest reliance among the five with the lowest source error. Over the $78$ test sets above, none of these rules reaches the mean regret of SRA, $2.2$ points, even with its threshold chosen on the outcomes of the test sets themselves. The best is the last rule, with $3.0$ points (difference from SRA $0.8$, $95\%$ CI $[0.1,1.7]$), and the switch to the fixed POPE choice has $3.3$ (difference $1.2$, $[0.1,2.4]$). With the threshold chosen on the other test-set types instead, the differences grow to $1.1$ and $2.5$ points. On the $68$ test sets with a context shift only, the switch to the fixed POPE choice with its best threshold matches or slightly beats SRA ($-0.3$, $[-0.6,0.0]$), because in this candidate pool the fixed POPE choice is the choice of SRA at a large shift. Under a context shift the choice of SRA moves in two steps, from InternVL3-14B to Qwen3-VL-2B at an estimated proportion of $.23$ and to Qwen2.5-VL-7B at $.35$, both set by the source measurements alone, and on the constructed test sets the text coordinate moves it to InternVL3-1B. No single threshold reproduces these moves. These comparisons are post hoc, and the test sets share pools, so the intervals, from a bootstrap over test sets ($B{=}10{,}000$), are optimistic.

\subsection{A detector fitted on source annotations only}\label{app:rsrc}
The detectors of \cref{sec:dhat} take their error rates from labeled items of the target domain's own corpus. A variant fits all its constants on COCO train2017 instead. It uses the detector of \cref{app:newcorpora} with its threshold set by the same rule on COCO items (\cref{app:procedure}), partners mined from COCO's annotations for every corpus, and the rates for each item type from the $1{,}200$-item COCO calibration set plus $600$ COCO items whose object and top five partners are all absent. No OpenImages or Objects365 annotation enters it. It applies only to questions about COCO categories, so it makes no estimate on the $19$ uncurated test sets where fewer than half the questions qualify, and SRA keeps selection by source error there. It is the variant used in \cref{app:organic}. On the test sets of \cref{tab:testsets} other than those of \cref{app:organic}, it never moved a selection where few questions mislead, and it underestimated every test set where many questions mislead, by $.08$ to $.60$. On Objects365 it moved all four selections on the misleading test sets to Qwen2.5-VL-7B (error $.303\to.181$) without any Objects365 annotation. On the misleading COCO test sets it moved all four, to the same model (error $.064\to.066$). On OpenImages it estimated the misleading test sets at $.20$ and $.36$ against a true $.80$ and moved one of the two selections (regret $.136\to.016$ on that test set).

\subsection{Label budget}\label{app:budget}
\begin{figure*}[t]
\centering
\includegraphics[width=0.92\textwidth]{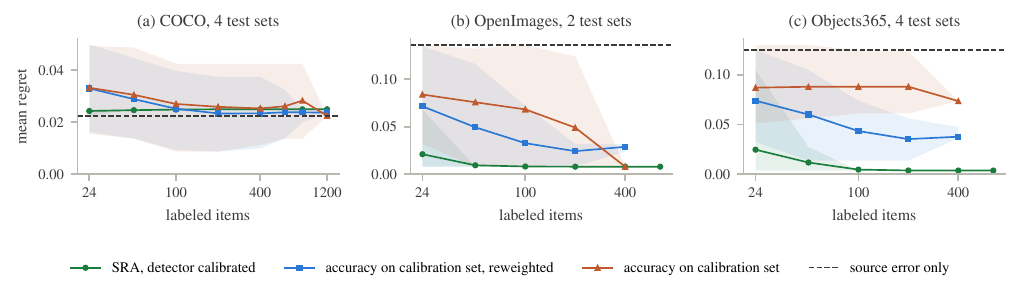}
\caption{\textbf{Selection regret against the number of labeled
items} on test sets where most questions mislead. Each
variant refits on a random subsample of the labeled items, $1{,}000$
draws per budget (one at the full budget), with bands from the 10th
to the 90th percentile.
The detector used on OpenImages and Objects365 also consumes
random absent items, counted in its budget, so its curve extends to
$650$. The dashed line is selection by source error, which uses no annotation. Ties are scored as the mean of the tied models. Reweighted accuracy also uses detector items that its budget does not count.}
\label{fig:budget}
\end{figure*}

\Cref{fig:budget} subsamples the labeled items of each corpus and reruns every variant's own fitting rule on the subsample. Draws keep the calibration set's class ratio. On COCO the calibrated detector refits its threshold, sensitivity, and false-positive rate on the subsample. Its regret stays at $.024$ to $.025$ at every budget, against $.023$ for source error and $.023$ to $.033$ for the variants that use accuracy on the calibration set, and neither accuracy variant beats source error at any budget up to $1{,}200$. On OpenImages and Objects365, SRA with the calibrated detector has lower regret than reweighted accuracy on the calibration set at every budget from $24$ items, and plain accuracy matches it only on OpenImages at the full $400$ items.

\paragraph{Labeled items drawn from the test sets.} The labeled items above come from each corpus's calibration set, where about $40\%$ of the items mislead on OpenImages and Objects365 and $50\%$ on COCO, against $80\%$ of the test questions, so plain accuracy compares the models at the wrong proportion and reweighting adds sampling noise. \Cref{tab:targetsample} instead draws the labeled items from the misleading test sets themselves, pooling the test sets of \cref{tab:testsets} and the new images of each corpus (cached answers). For each budget we draw the items without replacement at the test sets' own proportion, select the model with the lowest error on them, and score every selector's regret on the test sets with the drawn items removed, over $1{,}000$ draws. The SRA detector uses $650$ labeled items on OpenImages and Objects365, its calibration set and the random absent items of \cref{tab:protocol}, and $1{,}200$ on COCO, so the matched comparison is at those budgets. On OpenImages and Objects365, selection by labeled items trails SRA on average at every smaller budget and beats it in no draw, and at $650$ it ties SRA on OpenImages and trails it by $0.4$ points on Objects365, again without beating it in any draw. On COCO, where SRA and source error are tied, labeled items select better from $100$ items on, most often a Qwen3-VL model or LLaVA-OneVision-7B. Drawing the items from each test set alone, up to $100$ per set, gives the same picture.

\begin{table}[ht]
\centering\footnotesize
\caption{\textbf{Selection regret, in points, when the labeled items come from the test sets themselves}. $21$ candidates on the misleading test sets of each corpus, test sets of \cref{tab:testsets} and new images pooled. The labeled items are drawn at the test sets' proportion of misleading questions, and every selector is scored on the remaining items, $1{,}000$ draws per budget. Ties are scored as the mean of the tied models. The $650$ column is the detector's label budget on OpenImages and Objects365 (\cref{tab:protocol}). On COCO, whose detector uses $1{,}200$ items, labeled items have regret $2.4$ at that budget against $3.3$ for SRA and for source error.}
\label{tab:targetsample}
\setlength{\tabcolsep}{2pt}\scriptsize
\begin{tabular}{llcccccc}
\toprule
 & & \multicolumn{6}{c}{labeled items}\\
\cmidrule(lr){3-8}
corpus & selector & $24$ & $50$ & $100$ & $200$ & $400$ & $650$\\
\midrule
 & labeled items & $2.9$ & $2.5$ & $2.1$ & $1.8$ & $1.7$ & $1.8$\\
 & source error & $2.4$ & $2.4$ & $2.4$ & $2.4$ & $2.6$ & $2.7$\\
\oursrule
\multirow{-3}{*}{COCO} & \textbf{SRA (ours)} & $2.4$ & $2.4$ & $2.4$ & $2.4$ & $2.6$ & $2.7$\\
\midrule
 & labeled items & $5.1$ & $3.8$ & $2.7$ & $1.9$ & $1.5$ & $1.8$\\
 & source error & $13.0$ & $13.0$ & $13.1$ & $13.1$ & $13.3$ & $13.8$\\
\oursrule
\multirow{-3}{*}{OpenImages} & \textbf{SRA (ours)} & $1.0$ & $1.0$ & $1.0$ & $1.1$ & $1.3$ & $1.8$\\
\midrule
 & labeled items & $4.0$ & $2.6$ & $1.6$ & $1.1$ & $1.0$ & $1.3$\\
 & source error & $12.3$ & $12.3$ & $12.3$ & $12.4$ & $12.5$ & $12.8$\\
\oursrule
\multirow{-3}{*}{Objects365} & \textbf{SRA (ours)} & $0.4$ & $0.4$ & $0.5$ & $0.5$ & $0.7$ & $0.9$\\
\bottomrule
\end{tabular}
\end{table}

\subsection{Regret against the proportion of misleading questions}\label{app:regretfrac}
\Cref{fig:regimeaxis} pools the $21$ candidates' answers on each corpus's misleading test sets and its calibration set (all of it on COCO, its present items on the other two), $760$ present and $1{,}240$ misleading questions on COCO, $300$ and $200$ on OpenImages, and $410$ and $640$ on Objects365. For each proportion of misleading questions $d$ from $0$ to $1$ in steps of $.05$, we draw $400$ test sets of $200$ items with replacement and score each selector's regret. Present items answer yes and misleading questions no, so the yes proportion is $1-d$ and both ends hold a single label. The label-shift curve reweights source error on present and neutral absent source questions by that true balance. SRA uses the measured reliance, the source error of \cref{eq:sra}, and the true proportion. On COCO it switches away from the model selected by source error at $.25$, trails it by up to $7.2$ points at $.4$, and comes within $2.2$ points of the best model at $.8$. On OpenImages it stays within $0.6$ points of the best model from $.4$ on, and on Objects365 within $1.3$ points from $.7$ on. The intercept of \cref{eq:sra}, the balanced source error, assumes a target with as many present as absent questions. On a pool whose present proportion is $\pi$ it is off by $\tfrac12-\pi$ times the gap between the model's error on present objects and its false-yes rate on absent ones (\cref{rem:gap}). Reweighting the intercept to the true proportion, $1-d$ on the present and $d$ on the absent source questions, moves the COCO switch to $.35$ and the largest gap behind source error to $4.7$ points at $.5$, still before the best model changes, and over the $81$ test sets outside SCAM it changes five selections, two for the better and three for the worse, none where most questions mislead.

\subsection{Selection by test-set type}\label{app:calconf}
\Cref{tab:calconffull} gives the regret by test-set type for eight types. On the $29$ test sets where SRA changed the choice of source error, it lowered the chosen model's error on $27$, and on the other $58$ it kept the model selected by source error. The two exceptions are misleading COCO test sets, where the error rose by $.005$ and $.015$.

\begin{table*}[t]
\centering\scriptsize
\caption{\textbf{Selection by test-set type, eight types.} Mean regret, $21$ candidates, as in \cref{tab:calconf} without the fixed POPE choice, adding the mixed and POPE test sets and average confidence and the nuclear norm after Platt scaling. Raw nuclear norm, which runs every candidate, never does worse than selecting by source error in any column of \cref{tab:calconf}, but does on POPE. POPE scores are in sample for selecting by POPE error ($^{\ddagger}$). Colors as in \cref{tab:calconf}, with the random row as the red threshold and the constant row never green.}
\label{tab:calconffull}
\setlength{\tabcolsep}{1.7pt}
\begin{tabular}{lcccccccc}
\toprule
 & \multicolumn{5}{c}{test-set type (number of test sets)} & \multicolumn{3}{c}{misleading test sets}\\
\cmidrule(lr){2-6}\cmidrule(lr){7-9}
selector & uncurated ($32$) & constructed ($10$) & plausible ($6$) & mixed ($4$) & POPE ($3$) & COCO & OpenImages & Objects365\\
\midrule
source error & $.016$ & \cellcolor{failpastel}$.109$ & \cellcolor{failpastel}$.057$ & $.023$ & $.010$ & \cellcolor{ourspastel}$.023$ & \cellcolor{failpastel}$.136$ & \cellcolor{failpastel}$.125$\\
label-shift reweighted & $.016$ & \cellcolor{failpastel}$.109$ & \cellcolor{failpastel}$.057$ & $.023$ & $.010$ & \cellcolor{failpastel}$.049$ & \cellcolor{failpastel}$.212$ & \cellcolor{failpastel}$.130$\\
POPE error & \cellcolor{ourspastel}$.013$ & \cellcolor{failpastel}$.144$ & \cellcolor{failpastel}$.072$ & $.028$ & \cellcolor{ourspastel}$.005^{\ddagger}$ & $.034$ & \cellcolor{failpastel}$.212$ & \cellcolor{failpastel}$.134$\\
least answer bias & \cellcolor{failpastel}$.050$ & \cellcolor{failpastel}$.136$ & $.043$ & \cellcolor{failpastel}$.039$ & \cellcolor{failpastel}$.036$ & \cellcolor{failpastel}$.038$ & $.072$ & $.064$\\
random & $.043$ & $.102$ & $.050$ & $.035$ & $.028$ & $.037$ & $.111$ & $.087$\\
\oursrule
\textbf{SRA (ours)} & \cellcolor{notepastel}$.016$ & \cellcolor{ourspastel}$.019$ & \cellcolor{ourspastel}$.028$ & \cellcolor{ourspastel}$.019$ & \cellcolor{notepastel}$.010$ & \cellcolor{notepastel}$.025$ & \cellcolor{ourspastel}$.008$ & \cellcolor{ourspastel}$.004$\\
\midrule
\multicolumn{9}{l}{\emph{reference: runs every candidate on the target}}\\
avg.\ confidence, raw & $.019$ & $.071$ & $.018$ & \cellcolor{ourspastel}$.004$ & $.013$ & $.006$ & \cellcolor{failpastel}$.120$ & $.070$\\
avg.\ confidence, Platt & $.012$ & \cellcolor{failpastel}$.173$ & $.027$ & $.014$ & $.006$ & $.016$ & \cellcolor{failpastel}$.140$ & \cellcolor{failpastel}$.133$\\
nuclear norm, raw & $.015$ & \cellcolor{ourspastel}$.021$ & $.019$ & \cellcolor{ourspastel}$.004$ & $.012$ & $.011$ & \cellcolor{failpastel}$.120$ & $.083$\\
nuclear norm, Platt & \cellcolor{ourspastel}$.011$ & $.034$ & $.023$ & $.014$ & \cellcolor{ourspastel}$.002$ & $.020$ & \cellcolor{failpastel}$.168$ & \cellcolor{failpastel}$.128$\\
best of nine estimators & $.015$ & \cellcolor{ourspastel}$.021$ & \cellcolor{ourspastel}$.015$ & \cellcolor{ourspastel}$.004$ & $.004$ & \cellcolor{ourspastel}$.005$ & \cellcolor{failpastel}$.116$ & \cellcolor{ourspastel}$.051$\\
always Qwen3-VL-4B & $.022$ & \cellcolor{failpastel}$.165$ & $.018$ & $.008$ & $.016$ & $.005$ & \cellcolor{failpastel}$.120$ & $.070$\\
\bottomrule
\end{tabular}
\end{table*}

\subsection{Estimator baselines on eight test-set types}\label{app:lfall}
\begin{table*}[t]
\centering\footnotesize
\caption{\textbf{What each estimator and SRA variant uses.} Labeled source items, unlabeled target images read by the selector's own detector, candidate answers on target items, and labeled target items (the calibration set).}
\label{tab:inputs}
\setlength{\tabcolsep}{6pt}
\begin{tabular}{lcccc}
\toprule
selector & source labels & detector on target & run on target & target annot.\\
\midrule
source error & \checkmark & -- & -- & --\\
ATC, DoC & \checkmark & -- & all & --\\
COT & \checkmark & -- & all & yes/no balance\\
confidence-based rows of \cref{tab:lfall} & -- & -- & all & --\\
model agreement, majority vote & -- & -- & all & --\\
accuracy, calibration set & -- & -- & calibration set & \checkmark\\
\oursrule
\textbf{SRA, raw detector} & \checkmark, edits & \checkmark & -- & --\\
\textbf{SRA, detector calibrated} & \checkmark, edits & \checkmark & -- & detector only\\
\textbf{SRA, terms measured on the calibration set} & \checkmark & \checkmark & calibration set & \checkmark\\
\bottomrule
\end{tabular}
\end{table*}

\begin{table*}[t]
\centering\footnotesize
\caption{\textbf{Label-free selectors on eight test-set types},
$21$ candidates, mean regret of each selector's chosen model, with ties
scored as the mean of the tied models. Inputs as in \cref{tab:inputs}. The
SRA row repeats \cref{tab:calconf,tab:calconffull} ($21$ candidates) and the SRA row of \cref{tab:hc}. The last two columns are the misleading test sets of the new
corpora. The constant row, a reference rather than a selector, always selects Qwen3-VL-4B and is never colored green. The constant row is worse than a random selection on the constructed test sets ($.165$ against $.102$). Selecting by POPE error always picks InternVL3-8B, the model with the lowest error on POPE's three splits ($^{\ddagger}$ in sample). Green marks the best value in a column, red a regret above the random row's, and yellow an SRA regret that is neither. }
\label{tab:lfall}
\setlength{\tabcolsep}{4pt}
\begin{tabular}{lcccccccc}
\toprule
selector (regret $\downarrow$) & uncurated & constructed & plausible & mixed & POPE & COCO misleading & OpenImages & Objects365\\
\midrule
source error & $.016$ & \cellcolor{failpastel}$.109$ & \cellcolor{failpastel}$.057$ & $.023$ & $.010$ & $.023$ & \cellcolor{failpastel}$.136$ & \cellcolor{failpastel}$.125$\\
ATC & $.015$ & \cellcolor{failpastel}$.147$ & \cellcolor{failpastel}$.063$ & $.013$ & \cellcolor{ourspastel}$.004$ & $.023$ & \cellcolor{failpastel}$.120$ & \cellcolor{failpastel}$.098$\\
DoC & $.016$ & $.045$ & \cellcolor{failpastel}$.057$ & $.023$ & $.010$ & $.023$ & \cellcolor{failpastel}$.180$ & \cellcolor{failpastel}$.136$\\
average confidence & $.019$ & $.071$ & $.018$ & \cellcolor{ourspastel}$.004$ & $.013$ & $.006$ & \cellcolor{failpastel}$.120$ & $.070$\\
entropy & $.017$ & $.071$ & $.019$ & \cellcolor{ourspastel}$.004$ & $.013$ & \cellcolor{ourspastel}$.005$ & \cellcolor{failpastel}$.120$ & $.070$\\
MaNo, softmax only & $.017$ & $.071$ & \cellcolor{ourspastel}$.015$ & \cellcolor{ourspastel}$.004$ & $.013$ & \cellcolor{ourspastel}$.005$ & \cellcolor{failpastel}$.120$ & $.070$\\
MaNo, as published & $.022$ & \cellcolor{failpastel}$.175$ & $.032$ & $.008$ & $.016$ & \cellcolor{ourspastel}$.005$ & \cellcolor{failpastel}$.120$ & $.051$\\
nuclear norm & $.015$ & $.021$ & $.019$ & \cellcolor{ourspastel}$.004$ & $.012$ & $.011$ & \cellcolor{failpastel}$.120$ & $.083$\\
model agreement & $.017$ & $.056$ & $.024$ & \cellcolor{ourspastel}$.004$ & $.013$ & $.009$ & \cellcolor{failpastel}$.128$ & $.080$\\
majority-vote pseudo-labels & $.019$ & $.063$ & $.022$ & $.005$ & $.013$ & $.013$ & \cellcolor{failpastel}$.116$ & $.063$\\
\midrule
POPE error & \cellcolor{ourspastel}$.013$ & \cellcolor{failpastel}$.144$ & \cellcolor{failpastel}$.072$ & $.028$ & $.005^{\ddagger}$ & $.034$ & \cellcolor{failpastel}$.212$ & \cellcolor{failpastel}$.134$\\
\emph{always Qwen3-VL-4B} & $.022$ & \cellcolor{failpastel}$.165$ & $.018$ & $.008$ & $.016$ & $.005$ & \cellcolor{failpastel}$.120$ & $.070$\\
random & $.043$ & $.102$ & $.050$ & $.035$ & $.028$ & $.037$ & $.111$ & $.087$\\
\oursrule
\textbf{SRA (ours)} & \cellcolor{notepastel}$.016$ & \cellcolor{ourspastel}$.019$ & \cellcolor{notepastel}$.028$ & \cellcolor{notepastel}$.019$ & \cellcolor{notepastel}$.010$ & \cellcolor{notepastel}$.025$ & \cellcolor{ourspastel}$.008$ & \cellcolor{ourspastel}$.004$\\
\bottomrule
\end{tabular}
\end{table*}

\Cref{tab:lfall} scores the estimators of \cref{sec:setup}, from cached answers, with the estimators' inputs in \cref{tab:inputs}. MaNo appears twice. As published, with its softrun normalization of the logits~\cite{xie2024mano}, it selects worse than SRA on average, and with softmax normalization only it behaves like average confidence, entropy, and the nuclear norm~\cite{deng2023dispersity}. The three candidates with the lowest POPE error, InternVL3-8B, -1B, and -14B, trail the model with the lowest mean error over the six plausible test sets by $5.6$, $5.6$, and $4.1$ points. The confidence-based estimators make nearly the same choice everywhere. Average confidence, softmax MaNo, and the nuclear norm pick a Qwen3-VL model on $67$ of the $71$ unedited test sets, because these candidates give the most overconfident probabilities of answering yes. Platt scaling each candidate on its own source questions removes the fixed choice, raises the confidence estimators' regret on the test sets where most questions mislead, and lowers it on the uncurated and POPE test sets (\cref{tab:calconffull}). Model agreement and majority-vote pseudo-labels rank the models within a test set better than most confidence-based rows ($\bar\rho$ of $+.38$ to $+.90$ outside OpenImages).

\subsection{Target domains whose proportion of misleading questions we did not set}\label{app:organic}
Three target domains test compositions we did not set, with the COCO-only detector of \cref{app:rsrc}. The first domain holds every existence question that people wrote in VQAv2~\cite{goyal2017vqav2} validation about a COCO category, on images no other experiment uses ($492$ questions in two test sets). At least $9$ of $10$ human answers agree on each, and only $6.7\%$ of them are misleading questions, so people describing a scene rarely ask about an object it suggests but lacks. The second is the part of DASH-B~\cite{augustin2025dash} whose objects a COCO object detector can detect ($400$ questions in two test sets). Its web images were mined against PaliGemma and LLaVA-NeXT to trigger hallucination, and about half of its questions ask about a verified absent object. The third has four Objects365 test sets of other images with plausible questions (proportion of misleading questions $.51$--$.54$) and two of random questions, the Objects365 plausible row of \cref{tab:testsets}.

The detector estimates the proportion of misleading questions at $.19$--$.30$ on VQAv2, where it is $.07$, at $.22$--$.29$ on DASH-B, at $.38$--$.46$ on the plausible Objects365 test sets, and at $.05$--$.12$ on the ordinary ones, whose proportion we did not measure. It thus overestimates VQAv2's proportion and underestimates those of DASH-B and the plausible test sets. SRA changes one VQAv2 and one DASH-B selection and all four of those plausible Objects365 selections, with mean regret $.014$, $.143$, and $.015$. On DASH-B the cost of keeping InternVL3-14B is $.150$, because Idefics3-8B reaches error $.105$ against the selected model's $.255$, and the mined images mislead most candidates at once, so model agreement has regret $.155$ there. On the Objects365 test sets drawn by the plausibility rule, SRA ranks the models within test sets better than source error ($\bar\rho$ of $+.31$ against $+.16$).

\paragraph{DASH-B with a calibrated detector.} The detector above uses no DASH-B label. We also calibrated an OWLv2 detector (\cref{app:procedure}) on a labeled half of DASH-B (sensitivity $.976$, false-positive rate $.104$) and drew six disjoint test sets of $200$ items, half without the queried object, from the other half. The estimated proportion of misleading questions is $.47$ to $.52$ against a true $.50$, and SRA switches from InternVL3-14B to Qwen2.5-VL-7B on all six test sets, the model the fixed POPE choice also picks. Its mean regret is $.027$ against $.145$ for source error (difference $-.118$, $95\%$ CI $[-.146,-.092]$, $B{=}10{,}000$, items resampled within test sets with the detector's estimate recomputed). A protocol hashed in advance fixed the split, the detector, the test sets, the detector's estimates, and the choices of SRA before the $21$ candidates gave $12{,}201$ new answers on the images that contain the object. The answers on the images without the object predate it and decide the comparison, since on the new images InternVL3-14B has the lower error. The test therefore shows that a calibrated detector recovers DASH-B's proportion and moves SRA to the better of the two models, not that SRA finds DASH-B's best model. That model is Qwen3-VL-4B on four of the six test sets, and its SRA score exceeds that of Qwen2.5-VL-7B by less than $.0001$.

\subsection{Predicted against realized error}\label{app:abserror}
The SRA score serves to rank the candidates. In a post hoc analysis, \cref{tab:abserror} compares its level, read as a predicted error, with the realized error on the misleading test sets of \cref{tab:hc}, where $80\%$ of the questions mislead. Over the $21$ candidates, predicted and realized error correlate on every corpus, but the prediction is several times too high on COCO, and with audited labels it is also too high on OpenImages and Objects365. On these test sets the realized error of a model is its error on the present objects plus $.8$ times the extra error on the misleading questions, the two quantities that source error and reliance estimate (\cref{app:theory}). Neither estimate matches its counterpart in level on any corpus, but both follow the order of the candidates. Source error correlates with the error on present objects at $r=.71$ to $.85$ on COCO and Objects365 (on OpenImages nearly every present object is answered correctly), and reliance with the extra error at $r=.61$ to $.89$.

COCO's misleading questions are easy for nearly every candidate. About $70\%$ of them are answered correctly by all $21$, against $31\%$ on OpenImages and $35$ to $41\%$ on Objects365, and every candidate errs less on them than on either other corpus, also with audited labels. InternVL3-14B, whose context reliance exceeds that of Qwen2.5-VL-7B by $.21$, errs $3$ to $5$ points more often than it on COCO's misleading questions, against $15$ to $17$ elsewhere, and $15$ to $18$ points less often on COCO's present objects, where its source error is only $6$ points lower. Many COCO questions that count as misleading thus rarely cause errors, so the same proportion of misleading questions does not raise error equally in every corpus. The two effects cancel, so the two models tie on COCO. The detector is not the cause, since with the true proportion SRA makes the same choice, and in the split of \cref{eq:law} most of the overestimated gap between the two models comes from source error. Removing the items the label audit flagged changes the gaps between the two models by at most $1.3$ points. Whether exposure to COCO images in pretraining makes its misleading questions easier is untested.

\begin{table}[ht]
\centering\scriptsize
\caption{\textbf{Predicted against realized error} on the misleading test sets, mean over the $21$ candidates, in percent. MAE is the mean absolute difference between predicted and realized error, and $r$ the Pearson correlation over candidates. The audited column removes the items the label audit flagged, on the test sets of \cref{tab:testsets}, and MAE and $r$ use the original labels.}
\label{tab:abserror}
\setlength{\tabcolsep}{3pt}
\begin{tabular}{@{}llccccc@{}}
\toprule
corpus & test sets & predicted & realized & audited & MAE & $r$\\
\midrule
COCO & \cref{tab:testsets} & $37.0$ & $7.8$ & $6.9$ & $29.2$ & $.53$\\
 & new images & $35.2$ & $7.7$ & -- & $27.5$ & $.60$\\
OpenImages & \cref{tab:testsets} & $32.1$ & $28.7$ & $19.2$ & $4.6$ & $.52$\\
 & new images & $31.1$ & $28.3$ & -- & $3.8$ & $.56$\\
Objects365 & \cref{tab:testsets} & $31.3$ & $26.4$ & $20.4$ & $5.2$ & $.67$\\
 & new images & $27.1$ & $25.4$ & -- & $3.2$ & $.60$\\
\bottomrule
\end{tabular}
\end{table}

\section{Nine Held-Out Models}\label{app:dayzero}
Nine held-out models were never run on any test set or calibration set before their predictions were fixed. The models are LensVLM-9B~\cite{lensvlm2026}, MiMo-V2.6-Distill-Qwen-9B~\cite{mimo2026}, Intern-Decision-4B~\cite{interndecision2026}, Qwen3.5-9B, -4B, and -2B~\cite{qwen35_2026}, North-Micro-Vision-Instruct~\cite{northmicro2026}, LFM2.5-VL-3B~\cite{lfm25vl2026}, and Granite-Vision-4.1-4B~\cite{granitevision2026}. Six share the Qwen3.5 architecture. We measured each model on the $100$ source images only, recording source error, context reliance by removal, text reliance under four templates, and reliance on a pasted sticker label in two styles, in about five minutes per model. Their predictions for all $71$ unedited and $10$ constructed test sets and for DASH-B, with the estimated shift values, were fixed before any test set was run (about $26{,}500$ answers per model).

Reliance predicted the nine models' DASH-B false ``yes'' rates at Spearman $.80$ and their sticker failures at $.93$ (\cref{fig:dayzero}). With the source error over present objects that their predictions used, SRA had regret $.020$ against $.109$ for source error on the constructed test sets, on the misleading Objects365 test sets its selection had error $.279$ against $.303$, and on all $32$ uncurated test sets it made the same selection as source error. The per-model rank predictions on the uncurated test sets failed in one direction, with models of low source error ranking far worse than predicted (MiMo-V2.6-Distill-Qwen-9B, predicted $3$rd of $30$, realized $29$th) and models of high source error far better (Qwen3.5-9B, $18$th to $6$th), and the predicted order against InternVL3-14B on the misleading COCO test sets held for only four of the nine models. On those test sets a source error over present objects predicts error on present items at Spearman $+.86$ and error on absent items at $-.74$, so it cannot see failures on absent ones, which the source error of \cref{eq:sra} counts (next paragraph). The scripts, prediction files, and outputs are in the project repository~\todo{URL}.

\paragraph{Source error with all 30 candidates (retrospective).}\label{app:mixsrc} With the $30$ candidates and the shift estimates of this test, the source error of \cref{eq:sra} lowers the regret of SRA on the misleading OpenImages test sets from $.084$ to $.016$ and on Objects365 from $.113$ to $.015$, relative to the source error over present objects, and raises the mean Spearman correlation within each test set from $-.15$ to $+.48$ and from $-.16$ to $+.55$. It leaves regret on the constructed test sets at $.020$ but raises regret on the uncurated test sets from $.017$, the regret of the model source error selects, to $.038$ and on misleading COCO from $.020$ to $.034$. With the $18$ development models alone and the same shift estimates, the uncurated regret is $.030$. Where few questions mislead, a small estimated shift can move SRA away from the model selected by source error among models whose source errors are close.

\section{CueDial Details}\label{app:cuedial}
\paragraph{Families and items.} CueDial gathers the yes/no items of the test-set types of \cref{tab:testsets}, the DASH-B images of \cref{sec:transfer}, and the OpenImages and Objects365 calibration items into one table of $33{,}456$ questions, of which $32{,}621$ remain after the audit below. Each item records its image, object, label, and kind, which is present, absent, misleading through scene context (an absent object with at least two of its top-five co-occurrence partners present), misleading through text (a misleading overlay on a present object, or a question about a sticker or pasted word), or a DASH-B hard negative. The table stores the first-token probability of yes of all $40$ models on every item.

\paragraph{Sampling.} The eight families draw their misleading, neutral, and other-label items from the same source. The context families draw on COCO (the misleading, plausible, mixed, and uncurated COCO sets), LVIS, OpenImages, and Objects365 (including the Objects365 plausible sets), and the text families on the AMBER and COCO constructed sets, SCAM, and SynthSCAM. Test sets follow \cref{sec:cuedial}, sampled with a fixed seed. Items are split by a hash of their identifier into a public $70\%$ (test sets of $200$) and a hidden $30\%$ (test sets of $100$, half the $n$ of \cref{sec:cuedial}, and $88$ for context on OpenImages).

\paragraph{Selectors.} The selectors that never run a candidate on the target are source error and SRA. The selectors that run every candidate are ATC, DoC, average confidence, entropy, the nuclear norm, MaNo, and model agreement, defined as in \cref{app:lfall}, of which \cref{tab:cuedialhidden,tab:cuedial40} report five. Regret is scored as in \cref{sec:cuedial}, with ties as the mean of the tied models.

\begin{table*}[t]
\centering\scriptsize
\caption{\textbf{Public and hidden splits.} Worst-case mean regret $W_s$ over proportions of misleading questions, public split / hidden split, $30$ candidates, audited labels. SRA uses the proportion read by OCR on the text families and the true proportion on the context families. Green marks the lowest public $W_s$ in a row, red a $W_s$ more than $.10$ above that of source error, and yellow SRA where it is neither.}
\label{tab:cuedialhidden}
\setlength{\tabcolsep}{3pt}
\begin{tabular}{lccccccc}
\toprule
family & source error & \textbf{SRA} & ATC & DoC & avg.\ conf. & nuclear & agreement\\
\midrule
context, COCO & \cellcolor{ourspastel}$.022$/$.024$ & \cellcolor{notepastel}$.056$/$.065$ & $.034$/$.051$ & $.038$/$.044$ & $.031$/$.026$ & $.031$/$.025$ & $.025$/$.026$\\
context, LVIS & $.088$/$.115$ & \cellcolor{notepastel}$.053$/$.068$ & $.032$/$.059$ & $.030$/$.057$ & \cellcolor{ourspastel}$.021$/$.064$ & \cellcolor{ourspastel}$.021$/$.060$ & $.023$/$.067$\\
context, OpenImages & $.096$/$.081$ & \cellcolor{notepastel}$.063$/$.094$ & $.038$/$.050$ & \cellcolor{ourspastel}$.007$/$.030$ & $.094$/$.055$ & $.094$/$.050$ & $.091$/$.049$\\
context, Objects365 & $.045$/$.059$ & \cellcolor{notepastel}$.037$/$.055$ & $.066$/$.069$ & $.047$/$.065$ & $.040$/$.025$ & \cellcolor{ourspastel}$.033$/$.024$ & \cellcolor{ourspastel}$.033$/$.040$\\
text, AMBER & $.167$/$.124$ & \cellcolor{ourspastel}$.013$/$.028$ & \cellcolor{failpastel}$.406$/$.424$ & \cellcolor{failpastel}$.406$/$.423$ & \cellcolor{failpastel}$.315$/$.051$ & $.069$/$.036$ & $.172$/$.148$\\
text, COCO & $.259$/$.256$ & \cellcolor{ourspastel}$.027$/$.012$ & \cellcolor{failpastel}$.435$/$.418$ & \cellcolor{failpastel}$.434$/$.418$ & \cellcolor{failpastel}$.434$/$.378$ & $.053$/$.026$ & $.325$/$.296$\\
text, SCAM & $.046$/$.049$ & \cellcolor{notepastel}$.046$/$.049$ & $.034$/$.050$ & $.058$/$.032$ & $.040$/$.060$ & \cellcolor{ourspastel}$.023$/$.031$ & $.124$/$.126$\\
text, SynthSCAM & $.069$/$.065$ & \cellcolor{notepastel}$.069$/$.065$ & $.052$/$.050$ & $.049$/$.059$ & $.076$/$.091$ & \cellcolor{ourspastel}$.030$/$.045$ & \cellcolor{failpastel}$.183$/$.177$\\
\bottomrule
\end{tabular}
\end{table*}

\paragraph{Uncertainty (post hoc).} The test sets of a family share most of their items, so we bootstrap items within each family's pools and redraw the test sets with the rule above ($B{=}4{,}000$). With $30$ candidates, the $95\%$ intervals for the largest $W_s$ over the eight families are $[.061,.101]$ on the public split and $[.076,.150]$ on the hidden one for SRA, and $[.072,.134]$ and $[.052,.114]$ for the nuclear norm, and the paired differences, $[-.057,.012]$ public and $[-.010,.073]$ hidden, separate the two on neither split. The two selectors' mean $W_s$ over families tie on the public split, and on the hidden one that of SRA is higher by $.017$ ($[.006,.029]$). SRA's largest and mean $W_s$ are below those of source error, ATC, DoC, and agreement on both splits, with $30$ or $40$ models (intervals clear of zero), and below those of average confidence except on the hidden split with $30$ candidates, where the interval reaches zero. With all $40$ models, the mean $W_s$ of SRA is below that of the nuclear norm on the public split ($-.015$, $[-.024,-.005]$). The bootstrap holds the source measurements fixed.

\paragraph{All 40 models.}\label{app:cuedial40} The ten models outside the $30$ candidates are measured with the same code and source images as the $30$. \Cref{tab:cuedial40} scores the selectors over all $40$ models. SmolVLM2-500M's source error, $.47$, is among the highest of all $40$, so neither SRA nor selecting by source error picks it, and over all $40$ models the confidence-based selectors also fall behind on the sticker families.

\begin{table*}[t]
\centering\scriptsize
\caption{\textbf{CueDial over all $40$ models}, as in \cref{tab:cuedialhidden}. Green marks the lowest public $W_s$ in a row, red a $W_s$ more than $.10$ above that of source error, and yellow SRA where it is neither.}
\label{tab:cuedial40}
\setlength{\tabcolsep}{2.5pt}
\begin{tabular}{lccccccc}
\toprule
family & source error & \textbf{SRA} & ATC & DoC & avg.\ conf. & nuclear & agreement\\
\midrule
context, COCO & \cellcolor{ourspastel}$.022$/$.024$ & \cellcolor{notepastel}$.049$/$.049$ & $.035$/$.059$ & $.049$/$.046$ & $.063$/$.070$ & $.058$/$.068$ & $.025$/$.027$\\
context, LVIS & $.090$/$.118$ & \cellcolor{notepastel}$.043$/$.062$ & $.033$/$.063$ & $.041$/$.061$ & $.028$/$.062$ & $.028$/$.062$ & \cellcolor{ourspastel}$.027$/$.074$\\
context, OpenImages & $.096$/$.081$ & \cellcolor{notepastel}$.063$/$.095$ & \cellcolor{ourspastel}$.038$/$.047$ & $.066$/$.041$ & $.047$/$.047$ & $.046$/$.044$ & $.085$/$.078$\\
context, Objects365 & $.045$/$.060$ & \cellcolor{notepastel}$.056$/$.071$ & $.066$/$.073$ & $.085$/$.094$ & $.103$/$.085$ & $.081$/$.046$ & \cellcolor{ourspastel}$.036$/$.039$\\
text, AMBER & $.167$/$.124$ & \cellcolor{ourspastel}$.028$/$.028$ & \cellcolor{failpastel}$.406$/$.424$ & \cellcolor{failpastel}$.406$/$.423$ & \cellcolor{failpastel}$.317$/$.307$ & $.069$/$.061$ & $.186$/$.189$\\
text, COCO & $.259$/$.256$ & \cellcolor{ourspastel}$.027$/$.012$ & \cellcolor{failpastel}$.435$/$.418$ & \cellcolor{failpastel}$.434$/$.418$ & $.336$/$.374$ & $.060$/$.026$ & $.322$/$.325$\\
text, SCAM & \cellcolor{ourspastel}$.049$/$.050$ & \cellcolor{ourspastel}$.049$/$.050$ & $.121$/$.130$ & \cellcolor{failpastel}$.359$/$.274$ & \cellcolor{failpastel}$.395$/$.405$ & $.075$/$.059$ & \cellcolor{failpastel}$.169$/$.150$\\
text, SynthSCAM & \cellcolor{ourspastel}$.069$/$.066$ & \cellcolor{ourspastel}$.069$/$.066$ & \cellcolor{failpastel}$.344$/$.207$ & \cellcolor{failpastel}$.359$/$.339$ & \cellcolor{failpastel}$.406$/$.429$ & $.084$/$.085$ & \cellcolor{failpastel}$.208$/$.198$\\
\bottomrule
\end{tabular}
\end{table*}

\paragraph{Label audit.} Candidates for the audit were the items that at least $27$ of $30$ models answer against their label, in the context, ordinary, VQAv2, POPE, and uncurated test-set types, with $250$ random items of the same types as controls. They were shuffled into sheets of four images, each showing only an identifier and the question. An automated auditor, Claude Opus 5.5, outside the candidates and shown neither the label nor any answer, marked each object visible, absent, or unsure. A human checked $40$ of these items and $15$ POPE items, again shown only the image and the question, Among the items they could decide, they confirmed $24$ of $27$ flagged errors and all $8$ flagged POPE items, and found $1$ error among $14$ unflagged items. Several POPE errors are ambiguous rather than wrong, such as a dining table under plates, and the audit estimates that nearly one POPE question in ten has a wrong or ambiguous label (\cref{tab:cuedialaudit}), in line with the re-annotation of RePOPE~\cite{neuhaus2025repope}. After the flagged items are removed, SRA keeps its advantage on OpenImages and Objects365 and stays slightly behind source error on COCO (\cref{tab:auditedsel}), so these selections do not depend on the wrong labels.

\begin{table}[ht]
\centering\footnotesize
\caption{\textbf{Label audit by test-set type.} Items nearly all models answer against their label (audited), how many the auditor flags as wrong, and the estimated proportion of wrong labels in the type, which adds the control error rate of the type's audit on the items not audited ($.008$ for most types, $.054$ for POPE and uncurated). The auditor flags $722$ of the $837$ audited items as wrongly labeled and $107$ as unclear, against $6$ wrong of the $218$ random controls it could decide ($1$ of $126$ for most types, $5$ of $92$ for POPE and uncurated), and CueDial drops these $835$ items.}
\label{tab:cuedialaudit}
\setlength{\tabcolsep}{3pt}
\begin{tabular}{lrrrr}
\toprule
test-set type & items & audited & flagged wrong & est.\ wrong\\
\midrule
COCO misleading & $800$ & $8$ & $6$ & $1.5\%$\\
mixed & $800$ & $30$ & $29$ & $4.4\%$\\
Objects365 misleading & $800$ & $63$ & $54$ & $7.5\%$\\
Objects365 plausible & $1{,}200$ & $52$ & $41$ & $4.2\%$\\
Objects365 ordinary & $600$ & $2$ & $2$ & $1.1\%$\\
OpenImages misleading & $250$ & $31$ & $24$ & $10.3\%$\\
OpenImages ordinary & $360$ & $20$ & $16$ & $5.2\%$\\
plausible & $1{,}200$ & $110$ & $101$ & $9.1\%$\\
VQAv2 & $492$ & $1$ & $1$ & $1.0\%$\\
POPE & $9{,}000$ & $340$ & $296$ & $8.5\%$\\
uncurated & $6{,}400$ & $180$ & $152$ & $7.7\%$\\
\bottomrule
\end{tabular}
\end{table}

\begin{table}[ht]
\centering\footnotesize
\caption{\textbf{Selections on the misleading test sets with audited labels.} Error of each selector's choice on the misleading test sets of \cref{tab:hc}, $21$ candidates, with the original labels and after removing the items the audit flagged. The choices are those of \cref{tab:hc}, and every model's error falls, most on OpenImages. Green marks, per corpus, the lower error with the original labels, and yellow the SRA value where it is not the lower.}
\label{tab:auditedsel}
\setlength{\tabcolsep}{1.5pt}
\begin{tabular}{lccc}
\toprule
error $\downarrow$, original / audited & COCO & OpenImages & Objects365\\
\midrule
source error & \cellcolor{ourspastel}$.064$ / $.054$ & $.312$ / $.214$ & $.303$ / $.244$\\
\oursrule
\textbf{SRA, detector calibrated (ours)} & \cellcolor{notepastel}$.066$ / $.057$ & \cellcolor{ourspastel}$.184$ / $.091$ & \cellcolor{ourspastel}$.181$ / $.114$\\
SRA regret $\downarrow$ & $.025$ / $.024$ & $.008$ / $.014$ & $.004$ / $.005$\\
\bottomrule
\end{tabular}
\end{table}

\paragraph{Release.} We release item identifiers and download scripts rather than the original images, the edited images where their licenses allow and otherwise the code that regenerates them, the cached answers, the sampling and harness code, the audit decisions, and the labels of the public split, and keep the hidden labels for evaluation on request.

\section{Implementation and Reproducibility}\label{app:repro}

\subsection{Shift detectors}\label{app:procedure}
For an item asking about object $o$ with mined partners $p_1,\dots,p_5$, CLIP scores the prompts ``a photo of a $o$'', ``a photo with no $o$'', ``a photo'', and ``a photo of $p_1$ and $p_2$''. The uncorrected detector flags a misleading context question when the no-$o$ prompt beats the $o$ prompt and the partner prompt beats the plain one, and $\hat d_{\mathrm{ctx}}$ is the flagged fraction. The calibrated detector of \cref{sec:dhat} uses only the first comparison, with a threshold on its margin, because its sampling rule already requires the partners to be present. On COCO it has sensitivity $.377$ and false-positive rate $.213$ on the $1{,}200$-item calibration set. On OpenImages and Objects365 the absence call comes from a Faster R-CNN with COCO weights~\cite{ren2015fasterrcnn}, an item being called absent when its highest score for the queried category falls below a threshold fit on the OpenImages calibration set by maximizing Youden's index, and the second signal is the partner prompt beating the plain one. The COCO-only variant of \cref{app:rsrc} refits that threshold on the COCO calibration items by the same rule, and the DASH-B test of \cref{app:organic} uses OWLv2~\cite{minderer2023owlv2}, which covers DASH-B's objects, with the same rule fit on its calibration half. The OCR detector flags an item for $\hat d_{\mathrm{text}}$ when the recognized text contains the object token together with a negation word. No detector is an evaluated model, and none uses any test-set annotation. The shift measures how often a cue misleads, not how often it appears, so a common cue that agrees with the label adds nothing. COCO, Objects365, OpenImages, and VQAv2 images may occur in the candidates' pretraining, an exposure that every selector shares.

\subsection{Edits and controls}\label{app:artifacts}
Waterbirds images combine CUB birds~\cite{wah2011cub} with Places backgrounds~\cite{zhou2018places}. A background swap cuts the bird out with its CUB segmentation mask and pastes it onto another Waterbirds image of the target place type, whose own bird we remove by Telea inpainting~\cite{telea2004inpainting}. Each Waterbirds edit has a matched same-class control, namely a same-class background composite, an agreeing overlaid word, or an agreeing hint. On COCO we mask every labeled instance of the queried object with COCO's instance segmentation and inpaint the dilated mask with LaMa~\cite{suvorov2022lama}, so no annotated instance remains.

\paragraph{Traces of erased objects.} An erased object can leave traces, such as a shadow, that let a model answer yes for a reason other than context and inflate its measured context reliance. Running the COCO object detector of \cref{app:rsrc} on the $100$ edited source images finds the erased object with score at least $.3$ on $20$ of them. Dropping these items lowers mean context reliance over the $30$ models from $.274$ to $.228$ but leaves their ranking nearly unchanged (Spearman $.97$). It also strengthens every transfer correlation, from $.78$ to $.89$ on the raw COCO hallucination rate and from $.59$ to $.67$ on DASH-B. These traces add noise to the measurement, but they are not the source of its predictive signal. All other results use the unfiltered measurement.

\paragraph{Placebo edits.} We measured the effect of editing itself on two models (LLaVA-OneVision-0.5B and Qwen2-VL-2B, $99$ items), using placebo edits that inpaint object-sized background regions avoiding all labeled objects, twice per image. The average of the two placebo edits moves the log-odds by $0.32$--$0.41$ per model (root mean square over items, against the unedited image), and the two differ from each other by $0.14$--$0.18$, against $2.8$--$3.8$ for erasing the queried object. Erasing the object moves the answer to a question about the scene (``indoors?'') by only $0.17$--$0.26$, close to the placebo level.

\subsection{Implementation and compute}\label{app:impl}
The experiments collected more than a million yes/no answers across all test sets, calibration sets, and edited items, $238{,}000$ of them for the held-out models of \cref{app:dayzero} (per-test-set counts in the project repository~\todo{URL}).
The $21$ selection candidates are InternVL3 (1B, 2B, 8B, 14B)~\cite{zhu2025internvl3}, Qwen2-VL (2B, 7B)~\cite{wang2024qwen2vl}, Qwen2.5-VL (3B, 7B)~\cite{bai2025qwen25vl}, Qwen3-VL (2B, 4B, 8B)~\cite{bai2025qwen3vl}, LLaVA-1.5-7B~\cite{liu2024llava15}, LLaVA-1.6 (Vicuna-7B, Mistral-7B)~\cite{liu2024llavanext}, LLaVA-OneVision (0.5B, 7B)~\cite{li2024llavaonevision}, Idefics2-8B~\cite{laurencon2024idefics2}, Idefics3-8B~\cite{laurencon2024idefics3}, SmolVLM (500M, 2.2B), and SmolVLM2-2.2B~\cite{marafioti2025smolvlm}.
InternVL3-14B, Qwen3-VL-2B, and SmolVLM2-2.2B are the three held-out models of the selection tables, and the nine models of \cref{app:dayzero} are listed there. The Waterbirds study uses CLIP ViT-B/32 and
ViT-L/14~\cite{radford2021clip}, SigLIP base and
large~\cite{zhai2023siglip}, and SigLIP~2
base~\cite{tschannen2025siglip2} as contrastive models, and
LLaVA-OneVision-0.5B, Qwen2-VL-2B, Qwen2.5-VL-3B, LLaVA-1.5-7B, and
LLaVA-1.6-Mistral-7B as generative models. All models are loaded
through Hugging Face Transformers~\cite{wolf2020transformers}.

All experiments ran on a single $16$\,GB GPU, one model at a time. Contrastive models run in float32 and generative models in bfloat16, except those we loaded with 4-bit NormalFloat (NF4) quantization~\cite{dettmers2023qlora}. These are the 7B and 8B Qwen, OneVision, Idefics, and InternVL3 models, InternVL3-14B, Qwen3-VL-4B in the test-set evaluation, and six of the nine models of \cref{app:dayzero}. On COCO-style questions each answer is one forward pass, with the probability of ``yes'' read from the first answer token, and the question is ``Is there a [object] in the image? Answer yes or no.'' The Waterbirds generative models decode greedily with at most eight new tokens. No sampling is used, so repeated runs give identical answers. Measuring reliance for the COCO study takes about $1{,}200$ forward passes per model ($100$ source items under $9$ text conditions, $100$ removal items with their controls, and the $100$ present-object questions). That takes a few minutes per model and about $40$ minutes for InternVL3-14B in 4-bit. Evaluating one model on $6{,}400$ test questions takes $11$--$107$ minutes for the $18$ development models and up to about three hours for the largest held-out ones.

\end{document}